\documentclass[nohyperref]{article}

\usepackage{microtype}
\usepackage{graphicx}
\usepackage{booktabs}
\usepackage{hyperref}

\usepackage[accepted]{icml2022}
\usepackage{amsmath}
\usepackage{amssymb}
\usepackage{mathtools}
\usepackage{amsthm}
\usepackage{setspace}
\usepackage{adjustbox}
\usepackage[capitalize,noabbrev]{cleveref}
\usepackage{multirow}
\usepackage{subcaption}
\usepackage[titletoc,title]{appendix}
\usepackage{acronym}
\usepackage{nccmath}
\usepackage{psfrag}
\usepackage[hyphenbreaks]{breakurl}
\usepackage{dsfont}
\usepackage{amsmath,latexsym,amssymb,newlfont}

\theoremstyle{definition}
\newtheorem{teorema}{Theorem}

\newtheorem*{remark}{Remark}

\newcommand\V[1]  { \mathbf{#1} }
\newcommand\B[1]  { \boldsymbol{#1} }

\newcommand\set[1] {\mathcal{#1}}
\newcommand\up[1] {\mathrm{#1}}

\definecolor{applegreen}{rgb}{0.55, 0.71, 0.0}
\definecolor{ao}{rgb}{0.0, 0.5, 0.0}
\definecolor{red}{rgb}{1.0, 0.0, 0.0}
\definecolor{pg}{rgb}{0.96, 0.31, 0.28}
\definecolor{red(ncs)}{rgb}{0.77, 0.01, 0.2}
\definecolor{fireenginered}{rgb}{0.81, 0.09, 0.13}
\definecolor{opt}{rgb}{0.199, 0.382, 1}
\definecolor{lea}{rgb}{0, 0.46, 1}
\definecolor{tra}{rgb}{0.59, 0.76, 1}
\definecolor{pre}{rgb}{0, 0.87, 0.59}

\acrodef{SGD}[Nesterov's-SG]{subgradient method}
\acrodef{ASM}[ASM]{accelerated subgradient method}
\acrodef{AMRC}[AMRC]{adaptive minimax risk classifier}
\acrodef{MRC}[MRC]{minimax risk classifier}
\acrodef{ERM}[ERM]{empirical risk minimization}
\acrodef{RKHS}[RKHS]{reproducing kernel Hilbert space}
\acrodef{RRM}[RRM]{robust risk minimization}
\acrodef{DWM}[DWM]{dynamic weighted majority}
\acrodef{FOGD}[FOGD]{Fourier online gradient descent}
\acrodef{SP}[SP]{shifting perceptron}
\acrodef{RBP}[RBP]{randomized budget perceptron}
\acrodef{NOGD}[NOGD]{Nystr\"om online gradient descent}
\acrodef{OGD}[OGD]{online Gradient Descent}
\acrodef{RMSE}[RMSE]{root mean square error}
\acrodef{NORMA}[NORMA]{naive online regularized risk minimization algorithm}
\acrodef{RFF}[RFF]{random Fourier features}
\acrodef{MSE}[MSE]{mean squared error}
 
\icmltitlerunning{Minimax Classification under Concept Drift with Multidimensional Adaptation and Performance Guarantees}

\begin{document}

\twocolumn[
\icmltitle{Minimax Classification under Concept Drift with \\Multidimensional Adaptation and Performance Guarantees}

\begin{icmlauthorlist}
\icmlauthor{Ver\'onica \'Alvarez}{yyy}
\icmlauthor{Santiago Mazuelas}{yyy,zzz}
\icmlauthor{Jose A. Lozano}{yyy,aaa}
\end{icmlauthorlist}

\icmlaffiliation{yyy}{BCAM-Basque Center for Applied Mathematics, Bilbao, Spain}
\icmlaffiliation{zzz}{IKERBASQUE-Basque Foundation for Science}
 \icmlaffiliation{aaa}{Intelligent Systems Group, University of the Basque Country UPV/EHU, San Sebastián, Spain}

\icmlcorrespondingauthor{Ver\'onica \'Alvarez}{valvarez@bcamath.org}
\icmlcorrespondingauthor{Santiago Mazuelas}{smazuelas@bcamath.org}
\icmlcorrespondingauthor{Jose A. Lozano}{jlozano@bcamath.org}

\icmlkeywords{Supervised classification, concept drift, underlying distribution}

 \vskip 0.3in
]

\printAffiliationsAndNotice{}

\begin{abstract}
The statistical characteristics of instance-label pairs often change with time in practical scenarios of supervised classification. Conventional learning techniques adapt to such concept drift accounting for a scalar rate of change by means of a carefully chosen learning rate, forgetting factor, or window size. However, the time changes in common scenarios are multidimensional, i.e., different statistical characteristics often change in a different manner. This paper presents adaptive minimax risk classifiers (AMRCs) that account for multidimensional time changes by means of a multivariate and high-order tracking of the time-varying underlying distribution. In addition, differently from conventional techniques, \acsp{AMRC} can provide computable tight performance guarantees. Experiments on multiple benchmark datasets show the classification improvement of \acsp{AMRC} compared to the state-of-the-art and the reliability of the presented performance guarantees.
\end{abstract}

\section{Introduction}

The statistical characteristics describing the underlying distribution of instance-label pairs often change with time in practical scenarios of supervised classification {\cite{gama:2014, ditzler2015learning}}.  Such concept drift is common in multiple applications including electricity price prediction \cite{webb:2018}, spam mail filtering \cite{delany:2005}, and credit card fraud detection \cite{wang:2003mining}. For instance, in the problem of predicting electricity price increases/decreases, the statistical characteristics related to electricity demand, generation, and price often change over time due to varying habits and weather. Supervised classification in those scenarios is commonly referred to as learning under concept drift (e.g., \citealp{kolter:2007, webb:2018}), learning in a drifting (dynamic) scenario (e.g., \citealp{mohri2012new, shen:2019}), and online adaptive learning (e.g., \citealp{zhang2018adaptive, cutkosky2020parameter}). 

\begin{figure}
         \centering
                  \psfrag{data1abcdef}[l][l][0.7]{Class $y = 1$}
                  \psfrag{data2abcdef}[l][l][0.7]{Class $y = 2$}
                  \psfrag{data3abcdef}[l][l][0.7]{Rule}
                  \psfrag{t1}[][][0.7]{$\tau_1$}
                                    \psfrag{t2}[][][0.7]{$\tau_2$}
                                                      \psfrag{t3}[][][0.7]{$\tau_3$}
                                                                        \psfrag{t4}[][][0.7]{$\tau_4$}
                                                                          \psfrag{t5}[][][0.7]{\textcolor{gray}{$\tau_1$}}
                                                                            \psfrag{t6}[][][0.7]{\textcolor{gray}{$\tau_2$}}
                  \psfrag{-2}[][][0.5]{$-2$}
                                \psfrag{Time t1}[][][0.7]{Time $t_1$}
                                                                \psfrag{Time t2}[][][0.7]{Time $t_2$}
         \includegraphics[width=0.48\textwidth]{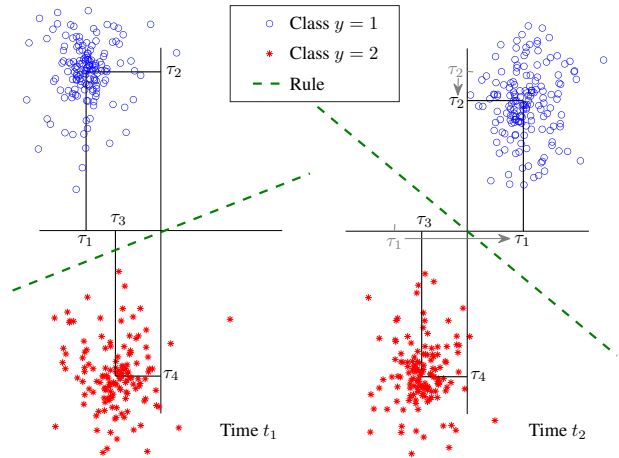}
\captionsetup{labelfont={it}, labelsep=period, font=small}
                                \caption{From time $t_1$ to time $t_2$, the statistical characteristic corresponding with class $1$ (${\tau}_1$ and ${\tau}_2$) significantly change, while those corresponding with class $2$ ($\tau_3$ and $\tau_4$) do not change. \acsp{AMRC} account for such multidimensional time changes by tightly tracking the time-varying underlying distribution.}               
                                \label{fig:changedistribution}
  \vskip -0.1in
\end{figure}

Supervised classification techniques adapt to concept drift by updating classification rules as new instance-label pairs arrive. Conventional learning techniques account for a scalar rate of change by means of a carefully chosen parameter such as a learning rate \cite{kivinen:2004, orabona:2008}, forgetting factor \cite{klinkenberg:2004, pavlidis:2011}, or window size \cite{bifet:2007, nguyen2018variational}. Specifically, a slow/fast rate of change is tackled by using a low/high learning rate, forgetting factor, or window size. More sophisticated techniques account for a time-varying scalar rate of change by adjusting the learning rates \cite{shen:2019}, the forgetting factors \cite{pavlidis:2011}, or the window sizes \cite{bifet:2007} over time. In particular, techniques based on dynamic regret minimization utilize a time-varying combination of rules obtained with different learning rates \cite{zhang2018adaptive, shen:2019}. 
However, in common scenarios, the concept drift cannot be adequately addressed accounting only for a scalar rate of change. Such inadequacy is due to the fact that time changes are commonly multidimensional, i.e., different statistical characteristics of instance-label pairs often change in a different manner (see Fig.~\ref{fig:changedistribution}). For instance, in the problem of predicting electricity price increases/decreases, the statistical characteristics related to demand often change differently from those related to generation. 

Conventional techniques based on statistical learning and empirical risk minimization do not provide performance guarantees under concept drift since instance-label pairs follow a time-varying underlying distribution. Techniques based on online learning and regret minimization provide performance guarantees for accumulated mistakes in terms of dynamic regret bounds \cite{kivinen:2004, cavallanti:2007, zhang2018adaptive, cutkosky2020parameter}. Furthermore, techniques based on statistical learning for the drifting scenario provide performance guarantees for the instantaneous error probability at specific times in terms of discrepancies between consecutive distributions \cite{long1999complexity, mohri2012new}. However, the existing techniques offer qualitative bounds but not computable tight performance guarantees. 

This paper presents \acfp{AMRC} that account for multidimensional time changes and provide tight performance guarantees. Specifically, the main contributions presented in the paper are as follows:
 \vspace{-0.1cm}
\begin{itemize}
   \setlength\itemsep{-0.1em}
\item We develop the learning methodology of \acp{AMRC} that provides multidimensional adaptation by estimating multiple statistical characteristics of the time-varying underlying distribution. 
\item We show that \acp{AMRC} can provide computable tight performance guarantees under concept drift in terms of instantaneous error probabilities and accumulated mistakes.
\item We propose techniques to track the time-varying underlying distribution using dynamical systems that model multivariate and high-order changes in statistical characteristics.
\item We propose techniques to learn \acp{AMRC} using an efficient subgradient method that utilizes warm-starts and maintains local approximations
of the objective function.
\item Using multiple benchmark datasets, we quantify the classification improvement of \acp{AMRC} in comparison with the state-of-the-art and numerically assess the reliability of the performance guarantees presented.
\end{itemize}
The rest of the paper is organized as follows. Section~\ref{sec:pre} briefly describes the problem formulation and \acp{MRC}. Section~\ref{sec:mrc} presents the learning methodology of \acp{AMRC} together with their performance guarantees. We propose sequential techniques for tracking and efficienct learning in Sections~\ref{sec:tr} and~\ref{sec:lear}. Section~\ref{sec:er} assesses the proposed methods using synthetic and benchmark datasets. 

\vspace{-0.2cm}
\paragraph{Notations:} calligraphic letters represent sets; bold lowercase letters represent vectors; bold uppercase letters represent matrices; $\V{I}$ denotes the identity matrix; $\mathds{1} \{\cdot\}$ denotes the indicator function; $\text{sign}(\cdot)$ denotes the vector given by the signs of the argument components; $\| \cdot \|_1$ and $\| \cdot\|_{\infty}$ denote the $1$-norm and the infinity norm of its argument, respectively; $\otimes$ and $\odot$ denote the Kronecker and Hadamard products, respectively; $(\,\cdot\,)_+$ and $[\,\cdot\,]^{\text{T}}$ denote the positive part and the transpose of its argument, respectively; $\preceq$ and $\succeq$ denote vector inequalities; $\mathbb{E}_{\up{p}}\{\,\cdot\,\}$ denotes the expectation of its argument with respect to distribution $\up{p}$; and $\V{e}_{i}$ denotes the $i$-th vector in a standard basis.

\section{Preliminaries} \label{sec:pre}
This section first recalls the basics of supervised classification in order to introduce the notation used and states the specific setting addressed. Then, we briefly describe \ac{MRC} methods that minimize the worst-case error probability over an uncertainty set.

\subsection{Problem formulation}
Supervised classification uses instance-label pairs to determine classification rules that assign labels to instances. We denote by $\mathcal{X}$ and $\mathcal{Y}$ the sets of instances and labels, respectively; both sets are taken to be finite with $\mathcal{Y}$ represented by $\{1, 2, ..., |\mathcal{Y}|\}$. We denote by $\text{T}(\mathcal{X}, \mathcal{Y})$ the set of all classification rules (both randomized and deterministic) and we denote by $\up{h}(y|x)$ the probability with which rule $\up{h} \in \text{T}(\mathcal{X}, \mathcal{Y})$ assigns label $y \in \mathcal{Y}$ to instance $x \in \mathcal{X}$ ($\up{h}(y|x) \in \{0, 1\}$ for deterministic classification rules). In addition, we denote by $\Delta(\mathcal{X}\times\mathcal{Y})$ the set of probability distributions on $\mathcal{X} \times \mathcal{Y}$ and by $\ell(\up{h}, \up{p})$ the expected 0-1 loss of  the classification rule $\up{h}\in \text{T}(\mathcal{X}, \mathcal{Y})$ with respect to distribution $\up{p} \in \Delta(\mathcal{X}\times\mathcal{Y})$. If $\up{p}^*\in \Delta(\set{X}\times\set{Y})$ is the underlying distribution of the instance-label pairs, then $\ell(\up{h},\up{p}^*)$ is the error probability of rule $\up{h}$ denoted in the following as $R(\up{h})$.
 
Instance-label pairs are usually represented as real vectors by using a feature mapping $\Phi: \set{X}\times \set{Y} \rightarrow \mathbb{R}^m$. The most common way to define such feature mapping is using multiple features over instances together with one-hot encodings of labels as follows (see e.g., \citealp{mohri:2018})
\begin{align}
\label{eq:feature}
 \Phi(x, y)  = & \, \V{e}_{y} \otimes {\Psi}(x)=  \big{[}\mathds{1}\{y = 1\} {\Psi}(x)^{\text{T}}, \\
 & \mathds{1}\{y = 2\} {\Psi}(x)^{\text{T}}, ..., \mathds{1}\{y = |\mathcal{Y}|\} {\Psi}(x)^{\text{T}}\big{]}^{\text{T}}\nonumber
\end{align}
 where the map $\Psi : \mathcal{X} \rightarrow \mathbb{R}^d$ represents instances as real vectors. For instance, such map can be given by \ac{RFF}, that is
\begin{align}
\label{eq:RRF}
\Psi(x) = \big{[}&\cos(\V{u}_1^{\text{T}} \V{x}), \cos(\V{u}_2^{\text{T}} \V{x}),..., \cos(\V{u}_D^{\text{T}} \V{x}), \\
&\sin(\V{u}_1^{\text{T}} \V{x}), \, \sin(\V{u}_2^{\text{T}} \V{x}), ..., \, \sin(\V{u}_D^{\text{T}} \V{x})\big{]}^{\text{T}} \nonumber
\end{align}
for $D$ random Gaussian vectors $\V{u}_1, \V{u}_2, ..., \V{u}_D \sim N(0, \gamma \V{I})$ with covariance given by the scaling parameter $\gamma$ (see e.g., \citealp{lu:2016, nguyen:2017, shen:2019}). 

In supervised classification under concept drift, samples arrive over time and the instance-label pair $(x_t,y_t)$ at time $t$ is a sample from an underlying distribution $\up{p}_t \in \Delta(\set{X} \times \set{Y})$ that changes over time. Learning methods use each new pair to update the previous classification rule. Specifically, at each time $t$, these techniques predict a label $\hat{y}_t$ corresponding with a new instance $x_t$ using the rule $\up{h}_t$ at time $t$, then they obtain updated rule $\up{h}_{t+1}$ when the true label $y_t$ is provided. 

\subsection{Minimax risk classifiers}
\acp{MRC} learn classification rules by minimizing the worst-case error probability over distributions in an uncertainty set \cite{mazuelas:2020, mazuelas2022, mazuelas2022a}. Such techniques are especially suitable for supervised classification under concept drift because \acp{MRC} determine the uncertainty sets by expectation estimates and do not require to use instance-label pairs from the same underlying distribution. Other \ac{RRM} methods utilize uncertainty sets determined by metrics such as Wasserstein distances \cite{shafieezadeh2019regularization} and f-divergences \cite{namkoong2017variance}. However, such techniques would require a possibly complex tracking of uncertainty sets.

\acp{MRC} learn classification rules that are solutions to
 \begin{equation}
 \label{eq:minmaxrisk}
\underset{\up{h} \in \text{T}(\mathcal{X}, \mathcal{Y})}{\min} \, \underset{\up{p} \in \mathcal{U}}{\max} \; \ell(\up{h}, \up{p})
\end{equation}
where $\set{U}$ is an uncertainty set of distributions. Such uncertainty set is determined by expectation estimates as
\begin{equation}
\label{eq:us}
\mathcal{U} = \{\up{p} \in \Delta (\mathcal{X} \times \mathcal{Y}) : \left|\mathbb{E}_{\up{p}}\{\Phi(x, y)\}  - \widehat{\B{\tau}} \right| \preceq \B{\lambda}\}
 \end{equation}
where  $| \, \cdot \, |$ denotes the vector formed by the absolute value of each component in the argument, $\widehat{\B{\tau}}$ denotes the vector of expectation estimates corresponding with the feature mapping $\Phi$, and $\B{\lambda} \succeq \V{0}$ is a confidence vector that accounts for inaccuracies in the estimate.

As described in \citealp{mazuelas:2020, mazuelas2022a} using the 0-1 loss $\ell$, the MRC rule $\up{h}$ solution of~\eqref{eq:minmaxrisk} is given by a linear combination of the feature mapping that is determined by a parameter $\B{\mu}^*$. Specifically,  the \ac{MRC} rule $\up{h}$ assigns label $\hat{y} \in \mathcal{Y}$ to instance $x \in \mathcal{X}$ with probability
\begin{equation}
\label{eq:prob}
\up{h}(\hat{y}|x) = \left\{\begin{matrix}\left(\Phi(x, \hat{y})^{\text{T}} \B{\mu}^*  -  \varphi(\B{\mu}^*) \right)_+ /c_{x} & \text{if} \, c_{x} \neq 0\\
1/|\mathcal{Y}| & \text{if} \, c_{x} = 0 \end{matrix}\right.
\end{equation}
\begin{align}
\label{eq:varphi}
\hspace{-0.2cm}\mbox{with}\hspace{0.3cm}\varphi(\B{\mu}^*) &= {\underset{x \in \mathcal{X}, \set{C} \subseteq \mathcal{Y}}{\max}} \Big{(}\sum_{y \in \set{C}}\Phi(x, y)^{\text{T}}\B{\mu}^* - 1\Big{)}/|\set{C}|\hspace{0.2cm}\\
c_{x}&=\sum_{y \in \mathcal{Y}} \left(\Phi(x, y)^{\text{T}} \B{\mu}^* - \varphi(\B{\mu}^*) \right)_+.\nonumber
\end{align}
 Note that a label that maximizes the probability in~\eqref{eq:prob} is given by ${\arg} \max_{y \in \set{Y}} \up{h}(y|x) = \arg \max_{y \in \set{Y}} {\Phi}(x,y)^{\text{T}} \B{\mu}^*$. The deterministic classification rule $\up{h}^\up{d}$ that assigns such label will be referred in the following as deterministic MRC.  
 
The vector parameter $\B{\mu}^*$ that determines the \ac{MRC} rule is obtained by solving the convex optimization problem \cite{mazuelas:2020, mazuelas2022a}
\begin{equation}
\label{eq:mrc}
\underset{\B{\mu}}{\min} \; 1 - \widehat{\B{\tau}}^{\text{T}} \B{\mu} + \varphi(\B{\mu}) + \B{\lambda}^{\text{T}} \left|\B{\mu}\right|.
\end{equation}
In addition, the minimum value of~\eqref{eq:mrc} equals the minimax risk value of~\eqref{eq:minmaxrisk} that is denoted in the following as $R(\set{U})$.
 
\section{Methodology of adaptive MRCs}\label{sec:mrc} \label{sec:rw}

\begin{figure}
         \centering
                  \psfrag{x1}[l][l][1]{$x_{t-1}$}
                  \psfrag{y1}[l][l][1]{$y_{t-1}$}
                  \psfrag{y2}[l][l][1]{$\hat{y}_{t-1}$}
                                    \psfrag{o}[c][c][0.9]{\textcolor{opt}{Optimization}}
                                     \psfrag{w}[l][l][0.9]{Alg.~\ref{alg:upmu}}
                                         \psfrag{n}[c][c][0.9]{\textcolor{tra}{Tracking}}
                                    \psfrag{p}[][][0.9]{\textcolor{applegreen}{Prediction}}
                                     \psfrag{g}[][][0.9]{\textcolor{pg}{Performance}}
                                       \psfrag{s}[][][0.9]{\textcolor{pg}{guarantees}}
                                        \psfrag{q}[][][0.9]{Alg.~\ref{alg:upmu}}
                                    \psfrag{t}[c][c][1]{Tracking}
                                    \psfrag{b}[c][c][0.9]{\textcolor{lea}{Learning}}
                                                        \psfrag{j}[][][0.9]{Alg.~\ref{alg:amrc}}
                                                         \psfrag{r}[][][0.9]{Alg.~\ref{alg:pred}}
                  \psfrag{h1}[l][l][1]{$\up{h}_{t-1}$}
                  \psfrag{R1}[l][l][0.8]{$R(\set{U}_{t-1})$}
                   \psfrag{R2}[l][l][0.8]{$R(\set{U}_{t})$}
                  \psfrag{h2}[l][l][1]{$\up{h}_{t}$}
                          \psfrag{U2}[l][l][1]{$\set{U}_{t}$}
                                  \psfrag{U1}[l][l][1]{$\set{U}_{t-1}$}
                                   \psfrag{aa}[l][l][1]{}
                                          \psfrag{t1}[][][1]{$\widehat{\B{\tau}}_{t-1}$}
                                           \psfrag{t2}[][][1]{$\widehat{\B{\tau}}_{t}$}
                                            \psfrag{l1}[][][1]{${\B{\lambda}}_{t-1}$}
                                             \psfrag{l2}[][][1]{${\B{\lambda}}_{t}$}
                                             \psfrag{m2}[][][1]{${\B{\mu}}_{t}$}
                                             \psfrag{m1}[][][1]{${\B{\mu}}_{t-1}$}
                                             \psfrag{Alg4}[l][l][0.8]{Alg.~\ref{alg:upmu}}
                                             \psfrag{i}[l][l][0.9]{Alg.~\ref{alg:esttau}}
                                             \psfrag{Eq6}[l][l][0.8]{Eq.~\eqref{eq:prob}}
         \includegraphics[width=0.48\textwidth]{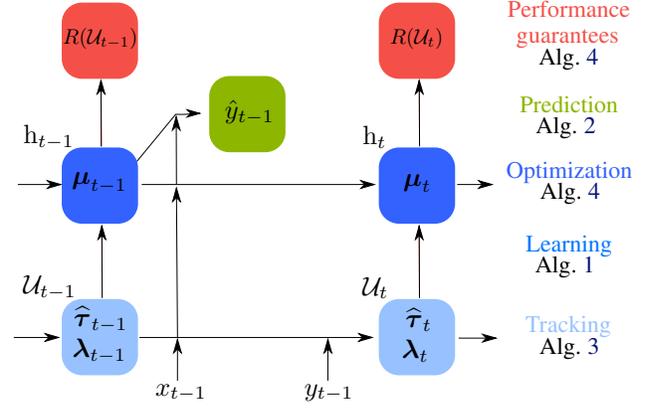}
\captionsetup{labelfont={it}, labelsep=period, font=small}
                                \caption{Diagram for the methodology of \acp{AMRC}.}                                                            
                                 \label{fig:diagram}
\end{figure}

This section first presents the learning methodology of \acp{AMRC}. Then, we describe how \acp{AMRC} provide multidimensional adaptation and tight performance guarantees.

Figure~\ref{fig:diagram} depicts the methodology of \acp{AMRC} that sequentially update \ac{MRC} rules as new instance-label pairs arrive, and Algorithms~\ref{alg:amrc} and~\ref{alg:pred} specify the learning and prediction stages of \acp{AMRC}. At learning, \acp{AMRC} adapt to multidimensional time changes by tightly tracking the time-varying underlying distribution (see Alg.~\ref{alg:esttau}); optimize the classifier parameters by using an efficient subgradient method (see Alg.~\ref{alg:upmu}); and obtain performance guarantees by using the minimax risk (see Alg.~\ref{alg:upmu} and Section~\ref{sec:pg}). 

The tracking step updates at each time the uncertainty set $\set{U}_t$ using the uncertainty set at previous time $\set{U}_{t-1}$. Specifically, the uncertainty set $\set{U}_t$ is determined by the mean and confidence vectors $\widehat{\B{\tau}}_t$ and $\B{\lambda}_t$ that are updated from those at time $t-1$ together with the instance-label pair $(x_{t-1}, y_{t-1})$. The optimization step updates at each time the classification rule $\up{h}_t$ and obtains the minimax risk $R(\set{U}_t)$ using the updated uncertainty set $\set{U}_t$ together with the rule at previous time $\up{h}_{t-1}$. Specifically, the classification rule $\up{h}_t$ is determined by parameter $\B{\mu}_t$ that is obtained from the updated mean and confidence vectors $\widehat{\B{\tau}}_t$ and $\B{\lambda}_t$ together with parameter $\B{\mu}_{t-1}$. 

\begin{algorithm}
\footnotesize
\captionsetup{labelfont={bf}, font=small}
\caption{Learning \acp{AMRC}}
\label{alg:amrc}
\begin{algorithmic}
\STATE \textbf{Input:} \hspace{0.3cm}$(x_{t-1}, y_{t-1})$, $\widehat{\B{\tau}}_{t-1}, \B{\lambda}_{t-1}$, and $\B{\mu}_{t-1}$
\STATE \textbf{Output:}\hspace{0.18cm}$\widehat{\B{\tau}}_{t}, \B{\lambda}_{t}$, $\B{\mu}_{t}$, and $R(\mathcal{U}_t)$
\STATE Update $\widehat{\B{\tau}}_t$ and ${\B{\lambda}}_t$ using $(x_{t-1}, y_{t-1})$, $\widehat{\B{\tau}}_{t-1}$, and ${\B{\lambda}}_{t-1}$ \\(see Alg.~\ref{alg:esttau})
\STATE Update $\B{\mu}_t$ and obtain $R(\mathcal{U}_t)$ solving~\eqref{eq:mrc} using $\widehat{\B{\tau}}_{t}, \B{\lambda}_{t}$, and $\B{\mu}_{t-1}$ (see Alg.~\ref{alg:upmu})
\end{algorithmic}
\end{algorithm}
\begin{algorithm}
\footnotesize
\captionsetup{labelfont={bf}, font=small}
\caption{Prediction with \acp{AMRC}}
\label{alg:pred}
\begin{algorithmic}
\STATE \textbf{Input:} \hspace{0.3cm}$x_t$ and $\B{\mu}_t$
\STATE\textbf{Output:}\hspace{0.18cm}$\hat{y}_t$ for AMRC $\up{h}_t$ or for deterministic AMRC $\up{h}_t^\up{d}$
\STATE$c_{x}\gets\sum_{y \in \mathcal{Y}} \left(\Phi(x_t, y)^{\text{T}} \B{\mu}_t - \varphi(\B{\mu}_t) \right)_+$
\STATE \textbf{if} {$c_x = 0$} \textbf{then }
\STATE \hspace{0.3cm}\textbf{for} {$y \in \mathcal{Y}$} \textbf{do }
\STATE \hspace{0.6cm}$\up{h}_t(y|x_t)\gets 1/|\set{Y}|$
\STATE \textbf{else}
\STATE \hspace{0.3cm}\textbf{for} {$y \in \mathcal{Y}$} \textbf{do }
\STATE \hspace{0.6cm}$\up{h}_t(y|x_t)\gets \left(\Phi(x_t, {y})^{\text{T}} \B{\mu}_t  -  \varphi(\B{\mu}_t) \right)_+ /c_{x}$
\STATE Draw $\hat{y}_t$ from distribution $\up{h}_t(y|x_t)$ or obtain $\hat{y}_t$ from $\arg \max_{y \in \set{Y}} \up{h}_t (y|x_t)$
		\end{algorithmic}
\end{algorithm}

\subsection{Multidimensional adaptation to time changes}\label{sec:mcd}
The proposed learning methodology can provide multidimensional adaptation to time changes because \acp{AMRC} estimate multiple statistical characteristics of the time-varying underlying distribution.

The proposed \acp{AMRC} estimate the evolution over time of the vector formed by the expectations of the feature mapping. At each time t, such mean vector ${\B{\tau}}_t~=~\mathbb{E}_{\up{p}_t}\{\Phi(x, y)\}~\in~\mathbb{R}^m$ represents the statistical characteristics of the underlying distribution $\up{p}_t$ as measured by the feature map $\Phi~:~\mathcal{X}~\times~\mathcal{Y}~\rightarrow~\mathbb{R}^m$. \acp{AMRC} account for situations in which different characteristics change in a different manner by estimating the evolution of each component in the mean vector.

For a feature mapping $\Phi$ given by one-hot encodings of labels as in~\eqref{eq:feature}, each of the $m=|\set{Y}|d$ components of the mean vector represents one conditional expectation. For $i=1,2, ..., m$, the $i$-th component $\Phi_i(x, y)$ has associated a label $j~=~1, 2, ..., |\mathcal{Y}|$ and an instance feature $\Psi_r$ for $r = 1, 2, ..., d$, since 
\begin{equation*}
\Phi_i(x, y) = \mathds{1}\{y = j\} {\Psi}_r(x)
\end{equation*}
with $i~=~(d-1)j+r$. Then, the $i$-th component of the mean vector $\B{\tau}_t$ corresponding to the $j$-th label and the $r$-th instance feature is given by 
 \begin{equation*}
\tau_{t, i} = \mathbb{E}_{\up{p}_t}\{\Phi_i(x, y)\} = \up{p}_{t}(y = j) \mathbb{E}_{\up{p}_t}\{\Psi_r(x)|y = j\}
\end{equation*}
where $\up{p}_{t}(y = j)$ denotes the probability of label $y = j$. The $i$-th component of the mean vector describes the probability of the $j$-th label together with the expected value of the $r$-th instance feature when the label takes its $\mbox{$j$-th}$ value. In Section~\ref{sec:tr} we propose techniques for estimating each of the mean vector components accounting for multivariate and high-order time changes.

\subsection{Performance guarantees}\label{sec:pg}
The proposed learning methodology can provide performance guarantees under concept drift because \acp{AMRC} minimize the worst-case error probability over distributions in a time-varying uncertainty set. 

The proposed \acp{AMRC} provide performance guarantees in terms of the minimax risks obtained at learning and also with respect to the smallest minimax risks corresponding with the optimal minimax rules. Such rules correspond with uncertainty sets given by the true expectation of the feature mapping $\Phi$, that is
\begin{equation}
\mathcal{U}_t^\infty = \{\up{p} \in \Delta (\mathcal{X} \times \mathcal{Y}) : \mathbb{E}_{\up{p}}\{\Phi(x, y)\} = {\B{\tau}}_t\}. 
 \end{equation}
 The minimum worst-case error probability over distributions in $\mathcal{U}_t^\infty$ is given by
\begin{equation}
\label{eq:smr}
R^\infty_t = \underset{\B{\mu}}{\min} \; 1 - {\B{\tau}}_{t}^{\text{T}} \B{\mu} + \varphi(\B{\mu}) = 1 - {\B{\tau}}_{t}^{\text{T}} \B{\mu}_t^\infty + \varphi(\B{\mu}_t^\infty)
\end{equation}
corresponding with the \ac{AMRC} given by parameters $\B{\mu}_t^\infty$. Such classification rule is referred to as optimal minimax rule because for any uncertainty set $\mathcal{U}_t$ given by~\eqref{eq:us} that contains the underlying distribution, we have that $\mathcal{U}_t^\infty \subseteq \mathcal{U}_t$ and hence $R^\infty_t \leq R(\mathcal{U}_t)$. This optimal minimax rule could only be obtained by an exact estimation of the mean vector that in turn would require an infinite amount of instance-label pairs at each time.

The following result shows performance guarantees for \acp{AMRC} both in terms of instantaneous bounds for error probabilities and bounds for accumulated mistakes.
\begin{teorema} \label{th:error_bounds}
For $t = 1, 2, ..., T$, let $\mathcal{U}_t$ be the uncertainty set given by $\widehat{\B{\tau}}=\widehat{\B{\tau}}_t$ and $\B{\lambda}= \B{\lambda}_t$ in~\eqref{eq:us}, and $\hat{y}_t$ be a label provided by an AMRC $\up{h}_t$ for $\mathcal{U}_t$ given by parameter $\B{\mu}_t$. Then, we have that
\begin{equation}
 \label{eq:bound2th}
R(\up{h}_t) \leq R(\mathcal{U}_t) + \alpha_t \leq R^\infty_t + \beta_t
\end{equation}
and with probability at least $1 - \delta$ 
\begin{align}
\label{eq:errorbound}
\sum_{t=1}^T \mathds{1} \left\{\hat{y}_t \neq y_t\right\} & \leq \sum_{t=1}^T R(\mathcal{U}_t) +\sum_{t=1}^T \alpha_t +  \sqrt{{2 T \log{\frac{1}{\delta}}}}\nonumber\\
&  \leq  \sum_{t=1}^T R^\infty_t +\sum_{t=1}^T \beta_t +  \sqrt{{2 T \log{\frac{1}{\delta}}}} 
\end{align}
where $\alpha_t$ and $\beta_t$ can be taken as 
\begin{align*}
\alpha_t & = \left\| |\B{\tau}_{t} - \widehat{\B{\tau}}_{t} | -\B{\lambda}_{t}\right\|_{\infty} \left\|\B{\mu}_t\right\|_1\\
 \beta_t & = (\left\|\B{\tau}_{t} - \widehat{\B{\tau}}_{t}\right\|_{\infty} + \left\|\B{\lambda}_{t}\right\|_{\infty})\left\|\B{\mu}_t^\infty - \B{\mu}_t\right\|_1
 \end{align*}
  for any $\B{\lambda}_t \succeq \V{0}$, and as 
  \begin{equation*}
  \alpha_t = 0, \, \beta_t = 2 \left\|\B{\lambda}_{t}\right\|_{\infty} \left\|\B{\mu}_t^\infty\right\|_1
  \end{equation*}
  for $\B{\lambda}_t \succeq \left|\B{\tau}_t - \widehat{\B{\tau}}_t\right|$.
  \begin{proof}
  See \ref{app:therror_bounds}.
  \end{proof}
\end{teorema}

Inequalities in~\eqref{eq:bound2th} and~\eqref{eq:errorbound} bound the instantaneous error probability and the accumulated mistakes of \acp{AMRC}, respectively. Inequalities in~\eqref{eq:bound2th} are obtained as a generalization of bounds in \citealp{mazuelas:2020, mazuelas2022a} to the addressed setting, while inequalities in~\eqref{eq:errorbound} are obtained using the  Azuma's inequality for the martingale difference $\mathds{1}\{\hat{y}_t \neq y_t\} - R(\up{h}_t)$. Note that the above inequalities also ensure generalization for deterministic \acp{AMRC} because $R(\up{h}_t^\up{d}) \leq 2 R(\up{h}_t)$ since $1 - \up{h}_t^\up{d}(y|x) \leq 2 (1 - \up{h}_t(y|x))$ for any $x \in \set{X}, y \in \set{Y}$. 

Existing methods based on statistical learning for the drifting scenario provide bounds for instantaneous error probabilities with respect to empirical risks (generalization bound) \cite{long1999complexity, mohri2012new}, while existing methods based on online learning and regret minimization provide bounds for accumulated mistakes with respect to that of comparator sequences (dynamic regret bound) \cite{kivinen:2004, cavallanti:2007, zhang2018adaptive, cutkosky2020parameter}. Theorem~\ref{th:error_bounds} for \ac{AMRC} methods provides bounds for instantaneous error probabilities and accumulated mistakes with respect to the smallest minimax risk $R^\infty_t$ corresponding to the optimal minimax rule (second inequalities in~\eqref{eq:bound2th} and~\eqref{eq:errorbound}).  

The proposed \acp{AMRC} not only provide qualitative bounds but also computable tight performance guarantees given by the minimax risk obtained at learning. However, existing methods provide bounds in terms of quantities that are not computable at learning such as discrepancies between consecutive distributions \cite{long1999complexity, mohri2012new} and comparators' path-length \cite{zhang2018adaptive, cutkosky2020parameter}. Theorem~\ref{th:error_bounds} for \ac{AMRC} methods provides computable tight bounds for instantaneous error probabilities and accumulated mistakes (first inequalities in~\eqref{eq:bound2th} and~\eqref{eq:errorbound}, respectively) that are given by the minimax risk obtained by solving optimization \eqref{eq:mrc} at learning. Specifically, the minimax risk $R(\set{U}_t)$ directly provides such bounds if the error in the mean vector estimate is not underestimated, i.e., $\B{\lambda}_t \succeq \left|\B{\tau}_t - \widehat{\B{\tau}}_t\right|$. In other cases, $R(\set{U}_t)$ still provides approximate bounds as long as the underestimation $|\B{\tau}_t - \widehat{\B{\tau}}_t| - \B{\lambda}_t$ is not substantial. In Section~\ref{sec:er} we show that the presented bounds can provide tight performance guarantees for AMRCs in practice. 

\section{Tracking the time-varying underlying distribution} \label{sec:tr}
This section describes the proposed algorithms for tightly tracking the time-varying underlying distribution. Such algorithms utilize methods that are commonly used for target tracking and describe target trajectories using kinematic models (see e.g., \citealp{bar:2004}). 

In what follows, we present techniques that estimate each component of the mean vector $\B{\tau}_t$ at time $t$ from instance-label pairs obtained up to time $t-1$. As described in Section~\ref{sec:mcd}, for $i = 1, 2, ..., m$,  we denote by $\tau_{t, i}$ the $i$-th component of the mean vector that corresponds with the $j$-th label and the $r$-th instance feature, that is, 
\begin{equation*}
\tau_{t, i} = \up{p}_{t}(y = j) \mathbb{E}_{\up{p}_t}\{\Psi_r(x)|y = j\}
\end{equation*}
 with $i = (d-1)j + r$ for $j = 1, 2, ..., |\mathcal{Y}|$ and $r = 1, 2, ..., d$. Then, the estimation of each component of the mean vector is obtained from the estimation of the corresponding label probability and instance feature conditional expectation. In particular, the $j$-th label probability for $j = 1, 2, ..., |\set{Y}|$ is estimated using the $W$ latest labels as 
 \begin{equation}
\label{eq:prob_num}
\mathrm{\widehat{p}}_{t}(y = j) = \frac{1}{W}\sum_{i = t-W}^{t-1} \mathds{1}\{y_i = j\}.
\end{equation}
The above conditional expectation denoted by $\gamma_{t, i}~=~\mathbb{E}_{\up{p}_t}\{\Psi_r(x)|y = j\}$ for $i = 1, 2, ..., m$ is recursively estimated as described in the following. 

We assume that $\gamma_{t, i}$ is $k$ times differentiable with respect to time and denote by $\B{\eta}_{t, i} \in \mathbb{R}^{k+1}$ the vector composed by $\gamma_{t, i}$ and its successive derivatives up to order $k$. As is usually done for target tracking \cite{bar:2004}, we model the evolution of such state vector $\B{\eta}_{t, i}$ using the partially-observed linear dynamical system 
\begin{equation}
\begin{split}
\label{eq:kalman}
\B{\eta}_{t, i} & = \V{H}_t \B{\eta}_{t-1, i} + \V{w}_{t, i}\\
\Phi_i(x_{t}, y_{t}) & = {\gamma}_{t, i} + v_{t, i}, \hspace{0.2cm} \text{if} \; y_t = j
\end{split}
\end{equation}
with transition matrix $\V{H}_t = \V{I} + \sum_{s=1}^k {\Delta_t^s}\V{U}_s/s!$
 where $\V{U}_s$ is the $(k+1)\times(k+1)$ matrix with ones on the $s$-th upper diagonal and zeros in the rest of components, $\Delta_t$ is the time increment at $t$, $j = 1, 2, ..., |\mathcal{Y}|$, and $i = (d-1)j+r$ for $r~=~1, 2, ..., d$.  The variables $\V{w}_{t, i}$ and $v_{t, i}$ represent uncorrelated noise processes with mean zero and variance  $\V{Q}_{t, i}$ and $r_{t, i}^2$, respectively. Such variances can be estimated online using methods such as those proposed in \citealp{odelson:2006, akhlaghi:2017}. Dynamical systems as that given by~\eqref{eq:kalman} are known in target tracking as kinematic state models and can be derived using the $k$-th order $\mbox{\text{Taylor}}$ expansion of ${\gamma}_{t, i}$ (see~\ref{app:taylor} for a detailed derivation).

The result below allows to recursively obtain state vector estimates $\widehat{\B{\eta}}_{t, i}$ with minimum \ac{MSE} together with their \ac{MSE} matrices $\widehat{\B{\Sigma}}_{t, i}$, for $i = 1, 2, ..., m$. Then, the first components of such vector and matrix provide the estimate and confidence for the conditional expectations $\gamma_{t, i}$  for $i = 1, 2, ..., m$.
\begin{teorema}\label{th:kalman}
If the evolution of the state vector $\B{\eta}_{t, i}$ is given by the dynamical system in~\eqref{eq:kalman}, then the linear estimator of $\B{\eta}_{t, i}$ based on $(x_1, y_1), (x_2, y_2), ..., (x_{t-1}, y_{t-1})$ that has the minimum \ac{MSE} is given by the recursion 
\begin{align}
\label{eq:eta_pred}
\widehat{\B{\eta}}_{t, i} & = \V{H}_{t} \widehat{\B{\eta}}_{t-1, i} - (\widehat{\B{\gamma}}_{t-1, i} - \Phi_i(x_{t-1}, y_{t-1})) \V{k}_{t, i}\\
\label{eq:sigma_pred}
\widehat{\B{\Sigma}}_{t, i} & = \V{H}_{t} \widehat{\B{\Sigma}}_{t-1, i} \V{H}_{t}^{\text{T}} + \V{Q}_{t, i} - \V{k}_{t, i} \V{e}_1^{\text{T}}\widehat{\B{\Sigma}}_{t-1, i} \V{H}_{t}^{\text{T}}
\end{align}
\begin{equation}
\label{eq:gain_vector}
\hspace{-0cm}\mbox{where}\hspace{0.2cm}\V{k}_{t, i} =  \mathds{1}\{y_{t-1} = j\}  \frac{ \V{H}_{t} \widehat{\B{\Sigma}}_{t-1, i} \V{e}_1}{\V{e}_1^{\text{T}} \widehat{\B{\Sigma}}_{t-1, i} \V{e}_1 + r_{t-1, i}^2}.
\end{equation}
In addition, $\widehat{\B{\Sigma}}_{t, i}$ is the \ac{MSE} matrix of such estimator $\widehat{\B{\eta}}_{t, i}$.
 \begin{proof} The unbiased linear estimator with minimum \ac{MSE} for a dynamical system such as~\eqref{eq:kalman} is given by the Kalman filter recursions (see e.g., \citealp{humpherys:2012}). Then, equations~\eqref{eq:eta_pred} and~\eqref{eq:sigma_pred} are obtained after some algebra from the Kalman recursions for predicted state vector and predicted \ac{MSE}.
\end{proof}
\end{teorema}

The above theorem enables to tightly track the time-varying underlying distribution. Specifically, Theorem~\ref{th:kalman} allows to recursively obtain mean vector estimates $\B{\widehat{\tau}}_t$ as well as its confidence vectors $\B{\lambda}_t$ every time a new instance-label pair is received. Such vectors are obtained from the estimated label probabilities in equation~\eqref{eq:prob_num} together with the estimated state vector and its \ac{MSE} in recursions~\eqref{eq:eta_pred} and~\eqref{eq:sigma_pred} as follows
\begin{align}
\label{eq:tauandlambda}
\widehat{{\tau}}_{t, i} & = \mathrm{\widehat{p}}_{t}(y = j) \, \V{e}_1^{\text{T}} \widehat{\B{\eta}}_{t, i} = \mathrm{\widehat{p}}_{t}(y = j) \, \widehat{{\gamma}}_{t, i}\\
\label{eq:lambda}
 {\lambda}_{t, i}^2 & = \mathrm{\widehat{p}}_{t}(y = j)(1-\mathrm{\widehat{p}}_{t}(y = j)) \V{e}_1^{\text{T}}\widehat{\B{\Sigma}}_{t, i}\V{e}_1 \nonumber\\
 +& \mathrm{\widehat{p}}_{t}(y = j)^2 \V{e}_1^{\text{T}}\widehat{\B{\Sigma}}_{t, i}\V{e}_1  + \widehat{{\gamma}}_{t, i}^2 \mathrm{\widehat{p}}_{t}(y = j)(1-\mathrm{\widehat{p}}_{t}(y = j)) \nonumber \\
 & = \mathrm{\widehat{p}}_{t}(y = j) \Big{(}\widehat{{\gamma}}_{t, i}^2 (1-\mathrm{\widehat{p}}_{t}(y = j)) + \V{e}_1^{\text{T}}\widehat{\B{\Sigma}}_{t, i}\V{e}_1\Big{)}
\end{align} 
with $i = (d-1)j+r$ for $j = 1, 2, ..., |\mathcal{Y}|$, $r = 1, 2, ..., d$.

\begin{remark}
The proposed techniques account for multivariate and high-order changes in statistical characteristics since the dynamical systems used model the detailed evolution of each component in the mean vector. Specifically, for $t = 1, 2, ...$ and $i= 1, 2..., m$, the $i$-th component of the mean vector estimate is updated at time $t$ by using the corresponding dynamical model in~\eqref{eq:kalman}, the estimate at previous time, and the most recent instance-label pair. Such update accounts for the specific evolution at time $t$ of the $i$-th component of the mean vector through the recursion in equation~\eqref{eq:eta_pred} and gain vector $\V{k}_{t, i}$ in equation~\eqref{eq:gain_vector}. In particular, updates for mean vector components and times with a low gain slightly change the estimate at previous time, while those updates for components and times with a high gain increase the relevance of the most recent instance-label pair.
\end{remark}

Algorithm~\ref{alg:esttau} details the proposed procedure to track the time-varying underlying distribution. Such algorithm has computational complexity $O(m k^3)$ and memory complexity $O(m k^2)$ where $m = |\set{Y}| d$ is the length of the feature mapping and $k$ is the order of the dynamical system in~\eqref{eq:kalman}. In Section~\ref{sec:er}, we use dynamical systems with orders $k = 0, 1, 2$ which are known in target tracking as zero-order derivative models, white noise acceleration models, and Wiener process acceleration models, respectively. 
    \begin{algorithm}
\footnotesize
\captionsetup{labelfont={bf}, font=small}
\caption{Tracking of the time-varying underlying distribution}
\label{alg:esttau}
    \setstretch{1.2}
\begin{algorithmic}
\STATE \textbf{Input:} \hspace{0.2cm}$(x_{t-1}, y_{t-1}), \widehat{\B{\eta}}_{t-1, i}, \widehat{\B{\Sigma}}_{t-1, i}, $ {$\widehat{\up{p}}_{t}(y = j), \V{Q}_{t, i}$}, and 
\STATE \hspace{1.15cm}$r_{t, i}$ for {$j = 1, 2, ..., |\mathcal{Y}|$} and $i = 1, 2, ..., m$
\STATE \textbf{Output:} $\widehat{\B{\tau}}_{t}, {\B{\lambda}}_{t}$, $\widehat{\B{\eta}}_{t, i}$, and $\widehat{\B{\Sigma}}_{t, i}$ for $i = 1, 2, ..., m$	
\STATE \textbf{for} {$i = 1, 2, ..., m$} \textbf{do }
\STATE \hspace{0.3cm}Obtain state vector $\widehat{\B{\eta}}_{t, i}$ using \eqref{eq:eta_pred} and \ac{MSE} $\widehat{\B{\Sigma}}_{t, i}$ using \eqref{eq:sigma_pred}
	\STATE \hspace{0.3cm}Obtain the $i$-th component of mean vector estimate $\widehat{\tau}_{t, i}$ 
	\STATE \hspace{0.3cm}using~\eqref{eq:tauandlambda} and confidence vector $\lambda_{t, i}$ using~\eqref{eq:lambda}
\end{algorithmic}
\end{algorithm}

\section{Efficient learning of \acp{AMRC}}\label{sec:lear}
This section describes the proposed algorithms that obtain \acp{AMRC}' parameters and minimax risks at each time. First, we describe the \ac{ASM} that solves the convex optimization problem~\eqref{eq:mrc} using Nesterov extrapolation strategy \cite{nesterov2015quasi, tao:2019}. Then, we propose efficient algorithms that use a warm-start for the \acs{ASM} iterations and maintain a local approximation of the polyhedral function $\varphi(\cdot)$ in~\eqref{eq:varphi}.

In what follows, we present techniques that efficiently obtain the classifier parameter $\B{\mu}_t$ and minimax risk $R(\set{U}_t)$ at time $t$ from the classifier parameter $\B{\mu}_{t-1}$ and the updated mean vector estimate $\widehat{\B{\tau}}_t$ and confidence vector $\B{\lambda}_t$. The \acs{ASM} algorithm applied to optimization~\eqref{eq:mrc} obtains classifier parameters using the iterations for {$l~=~1, 2, ..., K$}
\begin{align}
&\bar{\B{\mu}}_t^{(l+1)} = \B{\mu}_t^{(l)} + a_l \left(\widehat{\B{\tau}}_t - \partial \varphi({\B{\mu}_{t}^{(l)}}) - \B{\lambda}_t\odot\text{sign}(\B{\mu}_t^{(l)})\right) \nonumber\\
\label{eq:sgdrule}
&\B{\mu}_{t}^{(l+1)} = \bar{\B{\mu}}_t^{(l+1)} + \theta_{l+1}(\theta_{l}^{-1} - 1)\left({\B{\mu}}_t^{(l)} - \bar{\B{\mu}}_t^{(l)}\right)\end{align}
where $\B{\mu}_t^{(l)}$ is the $l$-th iterate for $\B{\mu}_t$, $\theta_l = 2/(l+1)$ and $a_l~=~1/(l+1)^{3/2}$ are step sizes, and $\partial \varphi(\B{\mu}_t^{(l)})$ denotes a subgradient of $\varphi(\cdot)$ at $\B{\mu}_t^{(l)}.$

The proposed algorithm reduces the number of \acs{ASM} iterations by using a warm-start that initializes the parameters {${\B{\mu}}_t$ in~\eqref{eq:sgdrule} with the solution obtained at previous time ${\B{\mu}}_{t-1}$}. In addition, the \acs{ASM} iterations are efficiently computed by maintaining a local approximation of the polyhedral function $\varphi(\cdot)$ in~\eqref{eq:varphi}. Such function is given by the pointwise maximum of linear functions indexed by pairs of instances and labels' subsets $x \in \mathcal{X}, \set{C} \subseteq \mathcal{Y}$. So that, if $\mathcal{I}$ is the set of such pairs we have that $\varphi(\B{\mu})$ in~\eqref{eq:varphi} becomes
\begin{equation}
\label{eq:maxvarphi}
\varphi(\B{\mu}) = \underset{i \in \mathcal{I}}{\max} \, \left\{\V{f}_i^{\text{T}} \B{\mu} - h_i\right\}
\end{equation}
with $\V{f}_i = {\sum}_{y \in \set{C}}\Phi(x, y)/|\set{C}| \in \mathbb{R}^m$ and $h_i = {1}/{|\set{C}|}$ for index $i$ that corresponds to pair $(x, \set{C})$.  
 We use local approximations of~\eqref{eq:maxvarphi} given by indices corresponding with {the $N$ most recent subgradients of $\varphi(\cdot)$}. Specifically, if $\set{R} \subseteq \mathbb{R}^m$ we have that for any $\B{\mu} \in \set{R}$, 
 \begin{equation*}
 \varphi(\B{\mu}) = \underset{i \in \mathcal{J}_{\set{R}}}{\max} \, \left\{\V{f}_i^{\text{T}} \B{\mu} - h_i\right\}
 \end{equation*}
  \begin{align*}
\hspace{-0.4cm}\mbox{with}\hspace{0.2cm} \mathcal{J}_{\set{R}} &= \{i \in \mathcal{I} : \exists \; \B{\mu} \in \set{R} \, \text{with} \, \varphi(\B{\mu}) = \V{f}_i^{\text{T}} \B{\mu} - h_i\} \\&= \{i \in \mathcal{I} : \exists \; \B{\mu} \in \set{R} \, \text{with} \, \V{f}_i \in \partial \varphi(\B{\mu})\}.
 \end{align*}
 Then, if $\mathcal{J}_{\set{R}} = \{i_1, i_2, ..., i_N\}$, we have that for any $\B{\mu} \in \set{R}$, $\varphi(\B{\mu}) = \max \{\V{F}\B{\mu} - \V{h}\}$ with $\V{F}~=~[\V{f}_{i_1}, \V{f}_{i_2}, ..., \V{f}_{i_N}]^{\text{T}}$ and $\V{h} = [h_{i_1}, h_{i_2}, ..., h_{i_N}]^{\text{T}}$.
 
 Algorithm~\ref{alg:upmu} details the proposed procedure to learn the classifier parameters $\B{\mu}_t$ together with the minimax risk $R(\mathcal{U}_t)$ that provides the performance guarantees. Such algorithm has computational complexity $O(N K m + m^2)$ and memory complexity $O(Nm)$ where $m = |\set{Y}| d$ is the length of the feature mapping, $N$ is the number of subgradients, and $K$ is the number of iterations for \acs{ASM} in~\eqref{eq:sgdrule}. Note that the complexity of Algorithms~\ref{alg:esttau} and~\ref{alg:upmu} does not depend on the number of instance-label pairs and time steps. Therefore, the proposed algorithms are applicable to large-scale datasets and have constant complexity per time step.

 \begin{algorithm}
\footnotesize
\setstretch{1.4}
\captionsetup{labelfont={bf}, font=small}
\caption{Optimization for ARMC params. and minimax risk}
\label{alg:upmu}
\begin{algorithmic}
\STATE \textbf{Input:} \hspace{0.3cm}$\widehat{\B{\tau}}_t$, ${\B{\lambda}}_t$, {${\B{\mu}}_{t-1}$}, $x_{t-1}$, $\V{F}$, and $\V{h}$
\STATE \textbf{Output:}\hspace{0.18cm}{${\B{\mu}}_{t}, R(\mathcal{U}_t)$, $\V{F}_{\text{new}}$, and $\V{h}_{\text{new}}$}
\STATE \textbf{for} {$\set{C} \subseteq \mathcal{Y}, \set{C}\neq\emptyset$} \textbf{do }
\STATE \hspace{0.3cm}$\V{F}, \V{h} \gets \text{append rows } \sum_{y \in \set{C}}\Phi(x_{t-1}, y)^{\text{T}}/|\set{C}|, 1/|\set{C}|\text{ to } \V{F}, \V{h}$
\STATE ${\B{\mu}}_{t}^{(1)} \gets {\B{\mu}}_{t-1}$, $\bar{\B{\mu}}_{t}^{(1)} \gets {\B{\mu}}_{t-1}$
\STATE \textbf{for }{$l = 1, 2, ..., K$} \textbf{do }
\STATE \hspace{0.3cm} {$a_l \gets 1/(l+1)^{3/2}$}, {$\theta_l \gets 2/(l+1)$, $\theta_{l+1} \gets 2/(l+2)$}
\STATE \hspace{0.3cm}$\V{f}_{i}^{\text{T}} \gets $ row of $\V{F}$ such that $\V{f}_i^{\text{T}} \B{\mu}_t^{(l)} - {h}_{i} =  {\max} \,\{ \V{F} \B{\mu}_t^{(l)} - \V{h}\}$
\STATE \hspace{0.3cm}$\bar{\B{\mu}}_t^{(l+1)} \gets \B{\mu}_t^{(l)} + a_l \left(\widehat{\B{\tau}}_t - \V{f}_i - \B{\lambda}_t \odot \text{sign} (\B{\mu}_t^{(l)})\right)$
\STATE \hspace{0.3cm}$\B{\mu}_{t}^{(l+1)} \gets \bar{\B{\mu}}_t^{(l+1)} + \theta_{l+1}(\theta_{l}^{-1} - 1)\left({\B{\mu}}_t^{(l)} - \bar{\B{\mu}}_t^{(l)}\right)$
\STATE\ \hspace{0.3cm}$\V{F}_{\text{new}}, \V{h}_{\text{new}} \gets \text{append rows } \V{f}_i^{\text{T}}, h_i \text{ to }  \V{F}_{\text{new}}, \V{h}_{\text{new}}$
				\STATE ${\B{\mu}}_{t} \gets \B{\mu}_{t}^{(K+1)}$
		\STATE $R(\mathcal{U}_t) \gets 1 -\widehat{\B{\tau}}_t^{\text{T}}\B{\mu}_t + \max \{\V{F} \B{\mu}_t - \V{h}\} + \B{\lambda}_t^{\text{T}}|\B{\mu}_t|$
				\STATE {$\V{F}_{\text{new}}, \V{h}_{\text{new}} \gets $ take the $N$ most recent $\V{f}_i$, $h_i$} from $\V{F}_{\text{new}}$, $\V{h}_{\text{new}}$
\end{algorithmic}
\end{algorithm}

\section{Numerical Results}\label{sec:er}

This section evaluates the performance of AMRCs in comparison with the presented performance guarantees and the state-of-the-art. In the first set of numerical results, we use a synthetic dataset, while in the second set of numerical results, we use multiple benchmark datasets. The implementation of the proposed \acp{AMRC} is publicly available in Python and Matlab languages.\footnote{\url{https://github.com/MachineLearningBCAM/AMRC-for-concept-drift-ICML-2022}} In addition, the appendices provide the detailed description of the benchmark datasets, additional implementation details, and supplementary numerical results. 

                         We utilize a type of synthetic dataset that has been often used as benchmark for supervised classification under concept drift \cite{kumagai:2016, kumagai:2017}. At each time $t$, we generate a label $y_t \in \{1, 2\}$ according to a Bernoulli distribution with parameter $1/2$ and an instance given by
\begin{align*}
\V{x}_t = \Big{[}&4\cos\left(\pi\Big((\cos(\omega t)-3)/{2} + y_t\Big)\right) + \epsilon_1,\\& 4\sin \left(\pi\Big((\cos(\omega t)-3)/{2} + y_t\Big)\right) + \epsilon_2\Big{]}^{\text{T}}
\end{align*}
with $\epsilon_1, \epsilon_2 \sim N(0, 2)$ and $\omega = 0.1$. In this paper we use a sinusoidal argument for cosines and sines so that the dataset is even more challenging. The class-conditional underlying distributions $\up{p}_t(\V{x}_t|y_t = 1)$ and $\up{p}_t(\V{x}_t|y_t = 2)$ are Gaussian with means that move with varying velocity and direction in a circle centered at the origin.

\acp{AMRC} are compared with the state-of-the-art using 12 datasets that have been often used as benchmarks for supervised classification under concept drift \cite{pavlidis:2011, kumagai:2016, lu:2016, nguyen:2017, webb:2018}: ``Weather'', ``Elec2", ``Airlines", ``German", ``Chess", ``Usenet1", ``Usenet2", ``Email Spam", ``Credit card", ``Smart grid stability", ``Shuttle", and ``Poker". The last $2$ datasets are multi-class problems and the rest are binary (see further details in the supplementary materials). The benchmark datasets can be obtained from UCI repository and from the Massive On-line Analysis library {\cite{bifet:2010}}. 

                           \begin{figure}
         \centering
                  \psfrag{Time}[c][c][0.8]{Time $t$}
\psfrag{Error}[c][t][0.8]{Error probability}
\psfrag{Order 0 method k = 0abc}[l][l][0.8]{$R(\up{h}_t)$ order $k = 0$}
\psfrag{Order 1}[l][l][0.8]{Order 1}
\psfrag{Order 2 method k = 2}[l][l][0.8]{$R(\up{h}_t)$ order $k = 2$}
\psfrag{Order 1 method k = 2}[l][l][0.8]{$R(\up{h}_t)$ order $k = 1$}
\psfrag{Upper boundabcdef}[l][l][0.8]{$R(\mathcal{U}_t)$ order $k = 1$}
                  \psfrag{0}[][][0.5]{}
                  \psfrag{20}[][][0.5]{}
                  \psfrag{40}[][][0.5]{40}
                \psfrag{60}[][][0.5]{}
                \psfrag{80}[][][0.5]{80}
                \psfrag{100}[][][0.5]{}
                \psfrag{120}[][][0.5]{120}
                \psfrag{140}[][][0.5]{}
                \psfrag{0.15}[][][0.5]{0.15}
                \psfrag{0.25}[][][0.5]{0.25}
                \psfrag{0.35}[][][0.5]{0.35}
                \psfrag{0.45}[][][0.5]{0.45}
                \psfrag{0.2}[][][0.5]{0.2}
                \psfrag{0.3}[][][0.5]{0.3}
                \psfrag{0.4}[][][0.5]{0.4}
                \psfrag{0.5}[][][0.5]{0.5}
  \includegraphics[width=0.45\textwidth]{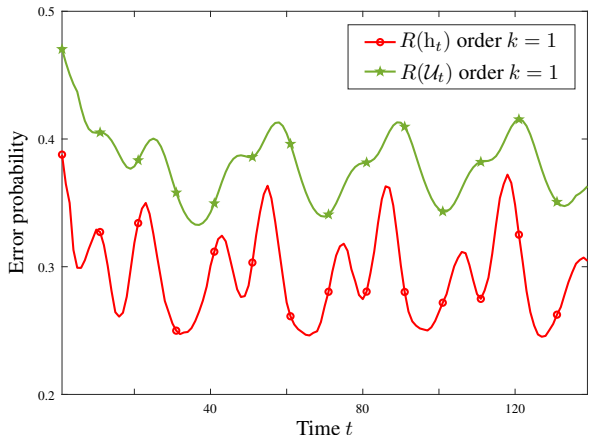}
   \vskip -0.1in
    \captionsetup{labelfont={it}, labelsep=period, font=small}
                     \caption{Results on synthetic data show the evolution in time of instantaneous performance bounds and error probabilities.}
         \label{fig:upper1}
              \end{figure}

                                                                           \begin{figure} 
                                                                           \centering
                                                             \psfrag{Mistakes rate order 0ab}[l][l][0.8]{AMRC order $k = 0$}
\psfrag{data2}[l][l][0.8]{AMRC order $k = 1$}
\psfrag{data3}[l][l][0.8]{AMRC order $k = 2$}
                  \psfrag{Time}[c][c][0.8]{Time $t$}
\psfrag{Mistakes bound}[l][l][0.8]{$\sum_{t=1}^{T} R(\mathcal{U}_t)/T +  \sqrt{{2\log{\frac{1}{\delta}}}/T}$ order $k = 1$}
                  \psfrag{Error}[c][t][0.8]{Accumulated mistakes per time}
                  \psfrag{0}[][][0.5]{0}
                                    \psfrag{10}[][][0.5]{10}
                                                                        \psfrag{100}[][][0.5]{100}
                                                                                                            \psfrag{1000}[][][0.5]{1000}
                                                                                                              \psfrag{10000}[][][0.5]{10000}
                \psfrag{0.26}[][][0.5]{0.26}
                \psfrag{0.34}[][][0.5]{0.34}
                \psfrag{0.42}[][][0.5]{0.42}
                \psfrag{0.44}[][][0.5]{0.44}
                \psfrag{0.5}[][][0.5]{0.5}
  \includegraphics[width=0.45\textwidth]{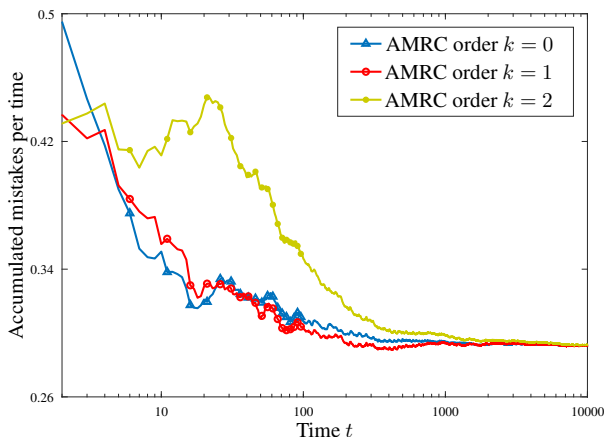}
          \vskip -0.1in
         \captionsetup{labelfont={it}, labelsep=period, font=small}
                \caption{Results on synthetic data show the evolution in time of accumulated mistakes per number of steps.}
                         \label{fig:syn_mistakes22}
  \vskip -0.2in
\end{figure}

\begin{table*}
 \captionsetup{labelfont={it}, labelsep=period, font=small}
                         \caption{Classification error in $\%$ of AMRCs in comparison with state-of-the-art techniques.}
    \vskip -0.15in
     \label{tb:results}
\setstretch{1.2}
\begin{center}
\scalebox{0.78}{\begin{tabular}{|c|r|r|r|r|r|r|r|r|r|r|r|r|r|}
\hline
\multicolumn{1}{|c|}{Algorithm} & \multicolumn{1}{c|}{Weather} & \multicolumn{1}{c|}{Elec2} & \multicolumn{1}{c|}{Airlines} & \multicolumn{1}{c|}{German} & \multicolumn{1}{c|}{Chess} & \multicolumn{1}{c|}{Usenet1} & \multicolumn{1}{c|}{Usenet2} & \multicolumn{1}{c|}{Email} & \multicolumn{1}{c|}{C. card} & \multicolumn{1}{c|}{S. Grid} & \multicolumn{1}{c|}{Shuttle} & \multicolumn{1}{c|}{Poker} & \multicolumn{1}{c|}{Ave. rank} \\ \hline
AdaRaker                      & 34.8                         & 49.5                       & 42.7                          & 41.0                        & 46.6                       & 50.0                         & 49.0                         & 45.7                       & 0.33                         & 46.5                         & N/A                          & N/A                        & 10.9                       \\ 
RBP                         & 34.3                         & 39.6                       & 42.7                          & 40.5                        & 36.5                       & 44.9                         & 48.3                         & 45.3                       & 0.33                         & 38.6                         & N/A                          & N/A                        & 8.0                      \\ 
Projectron                      & 30.7                     & 36.8                       & 41.8                          & 40.0                        & 35.1                       & 49.1                         & 47.7                         & 48.2                       & 0.33                         & \textbf{34.4}                     & 64.7                         & 22.6                       & 6.2                      \\
Projectron ++                       & 32.0                         & 48.1                       & 41.9                          & 42.2                        & 43.4                       & 47.2                         & 45.2                         & 48.1                       & 0.33                         & 35.8                     & 64.9                         & 23.4                       & 8.3                    \\ 
NORMA                         & 35.5                         & 43.2                       & 40.7                          & 39.7                        & 39.4                       & 36.0                         & 32.8                         & 34.6                   & 0.18                    & 38.8                         & N/A                          & N/A                        & 6.6                     \\ 
NOGD                         & 38.2                         & 48.2                       & 39.1                          & 37.1                        & 39.0                       & 34.6                    & 33.4                         & 36.1                       & 0.44                         & 38.7                         & 69.3                         & 23.4                       & 7.4                      \\ 
FOGD                         & 32.0                         & 44.9                       & 41.1                          & 37.6                        & 38.3                       & 45.9                         & 47.0                         & 44.9                       & 0.33                         & 36.0                     & 64.9                         & 22.9                       & 6.4                   \\ 
$\lambda$-perceptron                & 31.4                         & 40.4                       & 44.5                          & 42.9                        & 41.6                       & 44.5                         & 31.4                         & 46.8                       & 0.18                     & 35.8                    & N/A                          & N/A                        &7.3                       \\ 
DWM                         & \textbf{30.0}                    & 36.3                   & \textbf{37.8}                      & 41.2                        & 35.2                       & 36.3                         & \textbf{28.8}                     & 39.5                       & 0.80                         & 36.3                         & 16.9                     & 22.0                   & 4.8                        \\
Forgetron                      & 32.0                         & 46.0                       & 42.0                          & 40.1                        & 33.1                & 46.8                         & 47.2                         & 41.7                       & 0.33                         & 45.5                         & 64.4                         & 25.6                       &7.3                      \\  
Unidim. AMRC & 31.3 & 40.1          &       44.5                  &  30.3             &         38.3            &  46.3              &   33.3         &         48.4    &               0.18     &             36.2     &      70.4           &   39.4     &    7.2   \\ 
AMRC                          & 32.3                         & 35.8                  & 38.9                      & 30.3                    & \textbf{27.7}                   & 35.7                     & 30.9                     & 43.7                       & \textbf{0.17}                    & 35.8                    & 15.2                    & \textbf{21.9}                   & 3.0                      \\ 
Det.      AMRC                    & \textbf{30.0}                     & \textbf{33.9}                   & 39.4                          & \textbf{30.0}                   & 33.4                       & \textbf{32.0}                    & 29.9                         & \textbf{33.9}                   & \textbf{0.17}                     & \textbf{34.4}                     & \textbf{10.6}                     & \textbf{21.9}                   & \textbf{1.5}                     \\ 
 \hline
\end{tabular}}
\end{center}
   \vskip -0.25in
\end{table*}

The results for \acp{AMRC} are obtained using the feature mapping described in equation~\eqref{eq:feature} of Section~\ref{sec:pre}. Specifically, for the synthetic dataset we use the linear map $\Psi(x) = x$ and for the benchmark datasets we use \acp{RFF} with $D = 200$ as given by~\eqref{eq:RRF}. In addition, we use the recursive approach presented in \citealp{akhlaghi:2017} to obtain the variances {$\V{Q}_{t, i}$} and $r_{t, i}^2$ of noise processes $\V{w}_{t, i}$ and ${v}_{t, i}$, in~\eqref{eq:kalman}; we obtain the probability of the labels using~\eqref{eq:prob_num} with $W~=~200$; and the \ac{ASM} in~\eqref{eq:sgdrule} is implemented with $N = 100$ and $K~=~2000$.  \acp{AMRC} are compared with $10$ state-of-the-art techniques: AdaRaker \cite{shen:2019}, \ac{RBP} \cite{cavallanti:2007}, Projectron, Projectron ++ \cite{orabona:2008},  \ac{NORMA} \cite{kivinen:2004}, \ac{NOGD}, \ac{FOGD} \cite{lu:2016}, $\lambda$~-~perceptron \cite{pavlidis:2011}, \ac{DWM} \cite{kolter:2007}, and Forgetron \cite{dekel:2006}. The results of the 9 methods that use kernels utilize a scaling parameter calculated with a two-stage five-fold cross validation.

In the first set of numerical results we show the reliability of the presented instantaneous bounds using the synthetic data since the error probability in each time step cannot be computed using real-world datasets. Figure~\ref{fig:upper1} shows the averaged instantaneous bounds of error probabilities corresponding to first inequality in \eqref{eq:bound2th} for $\alpha_t = 0$ in comparison with the true error probabilities $R(\up{h}_t)$ at each time. Such figure shows that the instantaneous bounds given by $R(\mathcal{U}_t)$ can offer tight upper bounds for the error probability at each time. In addition, Figure~\ref{fig:syn_mistakes22} shows  the accumulated mistakes per time step of \acp{AMRC} of order $\mbox{k = 0, 1\text{, and }2}$. As can be seen in such figure, an increased order can result in an improved overall performance at the expenses of worse initial performance.

In the second set of numerical results we use benchmark datasets to quantify \acp{AMRC} performance with respect to the state-of-the-art, the improvement due to multidimensional adaptation, and the reliability of the presented mistake bounds. Table~\ref{tb:results} shows the classification error of \acp{AMRC} of order $k = 1$ and the $10$ state-of-the-art techniques for the 12 benchmark datasets described above.\footnote{Bold numbers indicate the top result and N/As indicates that the classification method cannot be used with non-binary datasets.} The table shows that \acp{AMRC} and deterministic \acp{AMRC} (Det. AMRC) achieve the best performance in 3 and 9 datasets, respectively, and are competitive in all datasets. In addition, the running time of \acp{AMRC} is of the order of tens of milliseconds per time step similarly to the state-of-the-art techniques (see further details in~\ref{app:imp_det}).

                                 \begin{figure}
                                  \centering
                                 \psfrag{data1}[l][l][0.8]{Deterministic AMRC}
                                                             \psfrag{AMRC RFFabcdefghijklmnop}[l][l][0.8]{AMRC}
\psfrag{Bound RFF}[l][l][0.8]{Bound AMRC}
\psfrag{Bound linear}[l][l][0.8]{Bound AMRC Linear}
\psfrag{AMRC linear}[l][l][0.8]{AMRC Linear}
\psfrag{DWM}[l][l][0.8]{DWM}
                  \psfrag{Time}[c][c][0.8]{Time $t$}
                  \psfrag{Error}[c][t][0.8]{Accumulated mistakes per time}
                  \psfrag{0.2}[][][0.5]{0.2}
                                    \psfrag{0.45}[][][0.5]{0.45}
                                                      \psfrag{0.7}[][][0.5]{0.7}
                                    \psfrag{0.95}[][][0.5]{0.95}
                                                                        \psfrag{5}[][][0.5]{5}
                                                                        \psfrag{0.45}[][][0.5]{0.45}
                \psfrag{0.5}[][][0.5]{0.5}
                \psfrag{0}[][][0.5]{0}
                \psfrag{250}[][][0.5]{250}
                \psfrag{500}[][][0.5]{500}
                \psfrag{750}[][][0.5]{750}
                \psfrag{1000}[][][0.5]{1000}
  \includegraphics[width=0.45\textwidth]{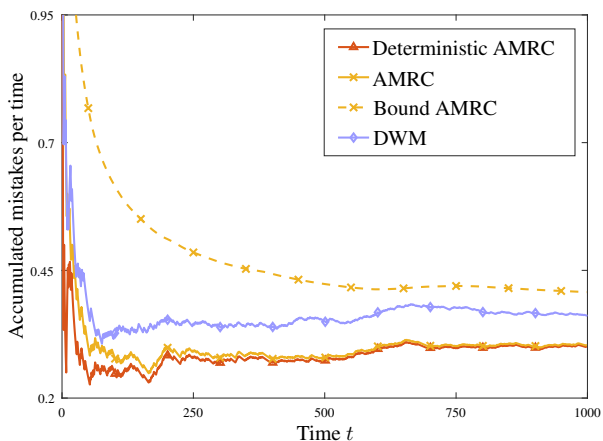}
  \captionsetup{labelfont={it}, font=small, }
 \caption{{Results on ``German'' dataset shows the evolution of accumulated mistake bounds and accumulated mistakes per number of steps.}}
                     \label{fig:german}
                     \vskip -0.1in
\end{figure}

                                 \begin{figure}
                                  \centering
                                   \psfrag{data1}[l][l][0.8]{Deterministic AMRC}
  \psfrag{AMRC RFFabcdefghijklmnop}[l][l][0.8]{AMRC}
\psfrag{Bound RFF}[l][l][0.8]{Bound AMRC}
\psfrag{Bound linear}[l][l][0.8]{Bound AMRC Linear}
\psfrag{AMRC linear}[l][l][0.8]{AMRC Linear}
\psfrag{DWM}[l][l][0.8]{DWM}
                  \psfrag{Time}[c][c][0.8]{Time $t$}
                  \psfrag{Error}[c][t][0.8]{Accumulated mistakes per time}
                  \psfrag{0.3}[][][0.5]{0.3}
                                    \psfrag{1}[][][0.5]{1}
                                                      \psfrag{0.7}[][][0.5]{0.7}
                                    \psfrag{10}[][][0.5]{10}
                                                                        \psfrag{1500}[][][0.5]{1500}
                                                                        \psfrag{0.4}[][][0.5]{0.4}
                \psfrag{0.1}[][][0.5]{0.1}
                \psfrag{0}[][][0.5]{0}
                \psfrag{250}[][][0.5]{250}
                \psfrag{500}[][][0.5]{500}
                \psfrag{750}[][][0.5]{750}
                \psfrag{1000}[][][0.5]{1000}
  \includegraphics[width=0.45\textwidth]{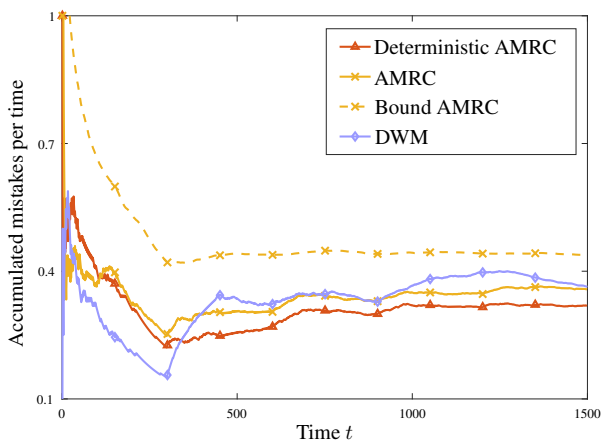}
\captionsetup{labelfont={it}, labelsep=period, font=small}
           \caption{Results on ``Usenet1'' dataset shows the evolution of accumulated mistake bounds and accumulated mistakes per number of steps.}
                     \label{fig:usenet1}
 \vskip -0.1in
\end{figure}

Table~\ref{tb:results} shows that \acp{AMRC} offer an overall improved performance compared to the state-of-the-art along the benchmark datasets. In order to more clearly assess the performance improvement due to the multidimensional adaptation, we also implement a simple version of deterministic \acp{AMRC} that only account for a scalar rate of change (Unidim. \acp{AMRC}). Such \acp{AMRC} are obtained by using, for all the components $i= 1, 2, ..., m$, the same gain in~\eqref{eq:eta_pred} and \eqref{eq:sigma_pred} that is taken as the average of the gains in~\eqref{eq:gain_vector}. As shown in the table, the presented multidimensional adaptation can enable significant performance improvements in most datasets.

Figures~\ref{fig:german} and \ref{fig:usenet1} provide more detailed comparisons using the $3$ techniques that have the top average rank (Det. \ac{AMRC}, \ac{AMRC}, and \ac{DWM}) in ``German" and ``Usenet1'' datasets. In addition, Figures~\ref{fig:german} and \ref{fig:usenet1} show the accumulated mistakes per number of steps $\sum_{t=1}^T \mathds{1} \left\{\hat{y}_t \neq y_t\right\}/T$ at each time in comparison with the accumulated mistake bounds per number of steps corresponding to the first inequality in~\eqref{eq:errorbound} for $\delta = 0.05$ and $\alpha_t = 0$. As can be seen in such figure, the accumulated mistake bounds in~\eqref{eq:errorbound} obtained at \ac{AMRC} learning can offer accurate estimates for the classification error. These results together with those in Figure~\ref{fig:upper1} show that the proposed methods can provide tight performance guarantees both in terms of instantaneous bounds for error probabilities and also in terms of bounds for accumulated mistakes. 

 \vspace{-0.35cm}
\section{Conclusion}\label{sec:conc}
 \vspace{-0.2cm}
The paper proposes the methodology of \acfp{AMRC} with multidimensional adaptation and performance guarantees. \acp{AMRC} learn classification rules estimating multiple statistical characteristics of the time-varying underlying distribution. In addition, \acp{AMRC} provide both qualitative and computable tight performance guarantees. The paper also proposes algorithms to track the time-varying underlying distribution accounting for multivariate and high-order time changes, and algorithms to efficiently update classification rules as new instance-label pairs arrive. The numerical results assess the performance improvement of \acp{AMRC} with respect to the state-of-the-art using benchmark datasets. The proposed methodology enables to more fully exploit data with characteristics that change over time and can provide more informative performance guarantees under concept drift.

\vspace{-0.35cm}
\section*{Acknowledgments}
 \vspace{-0.2cm}
Funding in direct support of this work has been provided by the Spanish Ministry of Science, Innovation and Universities through Ramon y Cajal Grant RYC-2016-19383, BCAM’s Severo Ochoa Excellence Accreditation SEV-2017-0718, and Project PID2019-105058GA-I00, the Spanish Ministry of Economic Affairs and Digital Transformation through Project IA4TES  MIA.2021.M04.0008, and by the Basque Government through the ELKARTEK, IT-1504-22, and BERC 2022-2025 programmes.

\newpage
\clearpage
\bibliography{bibliography}
\bibliographystyle{icml2022}

\newpage
\clearpage

\appendix
\gdef\thesection{Appendix \Alph{section}}
\section{Proof of Theorem~\ref{th:error_bounds}.} \label{app:therror_bounds}

\begin{proof}
For the first bound in~\eqref{eq:bound2th}, we have that $$R(\up{h}_t) \leq \underset{\up{p} \in \mathcal{U}_t^\infty}{\max} \; \ell (\up{h}_t, \up{p})$$ since $R(\up{h}_t) = \ell(\up{h}_t, \up{p}_t)$ and $\up{p}_t \in \mathcal{U}_t^\infty$ by definition of $\mathcal{U}_t^\infty$. Then, the maximization $$\underset{\up{p} \in \mathcal{U}_t^\infty}{\max} \; \ell (\up{h}_t, \up{p})$$ is obtained by using a slight reformulation of the optimization problem in the Theorem~2 in \citealp{mazuelas:2020}. Specifically, with the notation in this paper such optimization problem is 
\begin{equation*}
\begin{matrix}
\underset{\B{\mu}, \nu}{\min} & - {\B{\tau}}_t^{\text{T}} \B{\mu}  - \nu\\
\text{\, s. t. } & \Phi(x, y)^{\text{T}} \B{\mu} + \nu \leq \up{h}_t(y|x) - 1\\
&  \text{ for any } (x, y) \in \mathcal{X} \times \mathcal{Y} \end{matrix}
\end{equation*}
In addition, the constraint of the above optimization problem can be rewritten as
\begin{align*}
\Phi(x, y)^{\text{T}} & \B{\mu} + \nu \leq \up{h}_t(y|x) - 1,\ \forall x, y \\
& \Leftrightarrow  \nu  \leq \up{h}_t(y|x) - 1 - \Phi(x, y)^{\text{T}} \B{\mu},\ \forall x, y \\
 & \Leftrightarrow  \nu \leq - \underset{x \in \mathcal{X}, y \in \mathcal{Y}}{\max} \{\Phi(x, y)^{\text{T}} \B{\mu} - \up{h}_t(y|x) + 1\}.
\end{align*}
Then, minimizing $-\nu$, we obtain the optimization problem
\begin{align*}
\underset{\up{p} \in \mathcal{U}_t^\infty}{\max} & \; \ell (\up{h}_t, \up{p}) \\
& = \underset{\B{\mu}}{\min} \; 1 - {{\B{\tau}}}_t^{\text{T}} \B{\mu} + \underset{x \in \mathcal{X}, y \in \mathcal{Y}}{\max} \left\{ \Phi(x, y)^{\text{T}} \B{\mu} - \up{h}_t(y | x)\right\}
\end{align*}
that results in
\begin{align*}
\underset{\up{p} \in \mathcal{U}_t^\infty}{\max} & \; \ell (\up{h}_t, \up{p}) \\
& = \underset{\B{\mu}}{\min} \; 1 - {\B{\tau}}_t^{\text{T}} \B{\mu} + \underset{x \in \mathcal{X}, y \in \mathcal{Y}}{\max} \left\{ \Phi(x, y)^{\text{T}} \B{\mu} - \up{h}_t(y | x)\right\} \nonumber\\
& \leq 1 - {\B{\tau}}_t^{\text{T}} \B{\mu}_t + \underset{x \in \mathcal{X}, y \in \mathcal{Y}}{\max} \left\{ \Phi(x, y)^{\text{T}} \B{\mu}_t - \up{h}_t(y | x)\right\} \\
& \leq 1 - \B{\tau}_t^{\text{T}} \B{\mu}_t + \varphi(\B{\mu}_t)
\end{align*}
where the last inequality is due to the definition of $\up{h}_t$ and $\varphi(\cdot)$ in~\eqref{eq:prob} and~\eqref{eq:varphi}, respectively. Then, we have that
\begin{align}
& R(\up{h}_t) \leq 1 - \B{\tau}_t^{\text{T}} \B{\mu}_t + \varphi(\B{\mu}_t) \nonumber\\
& = 1 - \B{\tau}_t^{\text{T}} \B{\mu}_t + \varphi(\B{\mu}_t) + \widehat{\B{\tau}}_t^{\text{T}} \B{\mu}_t - \widehat{\B{\tau}}_t^{\text{T}} \B{\mu}_t + \B{\lambda}_t^{\text{T}} |\B{\mu}_t| - \B{\lambda}_t^{\text{T}}|\B{\mu}_t|  \nonumber\\
\label{eq:20}
& = R(\mathcal{U}_t) - \B{\tau}_t^{\text{T}} \B{\mu}_t + \widehat{\B{\tau}}_t^{\text{T}} \B{\mu}_t  - \B{\lambda}_t^{\text{T}}|\B{\mu}_t|
 \end{align}
 that leads to the first bound in~\eqref{eq:bound2th} for $\alpha_t~=~\left\| |\B{\tau}_{t} - \widehat{\B{\tau}}_{t} | -\B{\lambda}_{t}\right\|_{\infty} \left\|\B{\mu}_t\right\|_1$ using H\"{o}lder's inequality. In addition, the minimax risk given by the optimization problem~\eqref{eq:mrc} satisfies
\begin{align}
R(\mathcal{U}_t) & = \underset{\B{\mu}}{\min} \; 1 - \widehat{\B{\tau}}_{t}^{\text{T}} \B{\mu} + \varphi(\B{\mu}) + \B{\lambda}_{t}^{\text{T}} \left|\B{\mu}\right| \nonumber \\
& \leq 1 - \widehat{\B{\tau}}_t^{\text{T}} \B{\mu}_t^\infty + \varphi(\B{\mu}_t^\infty)  + \B{\lambda}_t^{\text{T}} \left|\B{\mu}_t^\infty\right| \nonumber\\
\label{eq:minmaxbound}
& = R^\infty_t + (\B{\tau}_t - \widehat{\B{\tau}}_t)^{\text{T}} \B{\mu}_t^\infty + \B{\lambda}_t^{\text{T}} \left|\B{\mu}_t^\infty\right|
\end{align}
that together with~\eqref{eq:20} leads to the second bound in~\eqref{eq:bound2th} {for $\beta_t = (\left\|\B{\tau}_{t} - \widehat{\B{\tau}}_{t}\right\|_{\infty} + \left\|\B{\lambda}_{t}\right\|_{\infty})\left\|\B{\mu}_t^\infty - \B{\mu}_t\right\|$} since 
\begin{align}
& R(\up{h}_t) \leq R(\mathcal{U}_t) - \B{\tau}_t^{\text{T}} \B{\mu}_t + \widehat{\B{\tau}}_t^{\text{T}} \B{\mu}_t  - \B{\lambda}_t^{\text{T}}|\B{\mu}_t| \nonumber \\
& \leq R^\infty_t + (\B{\tau}_t - \widehat{\B{\tau}}_t)^{\text{T}} (\B{\mu}_t^\infty-\B{\mu}_t) + \B{\lambda}_t^{\text{T}}(\left|\B{\mu}_t^\infty\right| - |\B{\mu}_t|)\nonumber\\
\label{eq:reverseti}
& \leq R^\infty_t + (\B{\tau}_t - \widehat{\B{\tau}}_t)^{\text{T}} (\B{\mu}_t^\infty-\B{\mu}_t) + \B{\lambda}_t^{\text{T}}\left|\B{\mu}_t^\infty - \B{\mu}_t\right|\\
& \leq R^\infty_t + \left(\left\|\B{\tau}_t - \widehat{\B{\tau}}_t \right\|_{\infty} + \left\|\B{\lambda}_t\right\|_{\infty}\right)\left\|\B{\mu}_t^\infty - \B{\mu}_t\right\|_1\nonumber
\end{align}
where~\eqref{eq:reverseti} is obtained using the reverse triangle inequality since $\B{\lambda}_t$ is positive.

For the first bound in~\eqref{eq:bound2th} in the case that $\B{\lambda}_t \succeq \left|\B{\tau}_t - \widehat{\B{\tau}}_t\right|$, we have that $R(\up{h}_t) \leq R(\mathcal{U}_t)$ because $\up{p}_t \in \mathcal{U}_t$ by definition of $\mathcal{U}_t$. Then, the minimax risk given by the optimization problem~\eqref{eq:mrc} satisfies~\eqref{eq:minmaxbound} that leads to~\eqref{eq:bound2th} for $\alpha_t = 0$ and $\beta_t = 2 \left\|\B{\lambda}_{t}\right\|_{\infty} \left\|\B{\mu}_t^\infty\right\|_1$ since $\B{\lambda}_t \succeq \left|\B{\tau}_t - \widehat{\B{\tau}}_t\right|$.

For the result in~\eqref{eq:errorbound}, let $(x_1, y_1), (x_2, y_2), ..., (x_T, y_T)$ be a sequence of instance-label pairs. If $$V_t~=~\mathds{1}\left\{\hat{y}_t\neq y_t\right\}~-~R(\up{h}_t)$$ we have that the sequence $V_1, V_2, ...$ is a martingale difference with respect to $(x_1, y_1), (x_2, y_2), ...$ because $$\mathbb{E}[V_t|(x_1, y_1), (x_2, y_2), ..., (x_{t-1}, y_{t-1})] = 0$$ for any $t$. Then, using the Azuma's inequality \cite{mohri:2018}, since $|V_t|\leq 1$ for any $t$, we have that
\begin{equation}
\label{eq:martdiff}
\sum_{t = 1}^T V_t \leq \sqrt{2 T \log \frac{1}{\delta}}
\end{equation}
with probability at least $1 - \delta$. Then, the result is obtained using bound in~\eqref{eq:bound2th} because with probability at least $1 - \delta$ we have that $$\sum_{t = 1}^T \mathds{1} \left\{\hat{y}_t \neq y_t\right\} \leq \sum_{t = 1}^T R(\up{h}_t) + \sqrt{2 T \log \frac{1}{\delta}}$$ using~\eqref{eq:martdiff} and the definition of $V_t$.
\end{proof}

\newpage
 \section{Derivarion of the dynamical system} \label{app:taylor}
 
Dynamical systems as that given by~\eqref{eq:kalman} can be derived using the $k$-th order Taylor expansion of the conditional expectations ${\gamma}_{t, i} = \mathbb{E}_{\up{p}_t}\{\Psi_r(x)|y = j\}$ with $i = (d-1)j + r$ for $j = 1, 2, ..., |\mathcal{Y}|$ and $r = 1, 2, ..., d$. The Taylor expansions of $\gamma_{t, i}$ and its successive derivatives up to order $k$ are given by
\begin{align*}
\gamma_{t, i} & \approx \gamma_{t-1, i} + \Delta_t \gamma_{t-1, i}' + \frac{\Delta_t^2}{2} \gamma_{t-1, i}'' + ... + \frac{\Delta_t^k}{k!} \gamma_{t-1, i}^{k)}\\
\gamma_{t, i}' & \approx \gamma_{t-1, i}' + \Delta_t \gamma_{t-1, i}'' + ... + \frac{\Delta_t^{k-1}}{(k-1)!} \gamma_{t-1, i}^{k)}\\
& \hspace{0.2cm} \vdots \\
\gamma_{t, i}^{k)} & \approx \gamma_{t-1, i}^{k)}
\end{align*}
where $\Delta_t$ is the time increment at $t$ and $\gamma_{t, i}^{k)}$ denotes the $k$-th derivative of $\gamma_{t, i}$. The above equations lead to
\begin{align}
\label{eq:e}
\B{\eta}_{t, i} &  \approx \begin{pmatrix} 1 & \Delta_t &  \frac{\Delta_t^2}{2} & ... &  \frac{\Delta_t^k}{k!} \\
0 & 1 & \Delta_t & ... &  \frac{\Delta_t^{k-1}}{(k-1)!}\\
\vdots & \vdots & \vdots &&\vdots\\
0 & 0 & 0 & ... & 1
\end{pmatrix} \B{\eta}_{t-1, i}
\end{align}
since $\B{\eta}_{t, i}$ is composed by $\gamma_{t, i}$ and its derivatives up to order $k$. In addition, $\gamma_{t, i}$ is observed at each time $t$ through the instance-label pair $(x_t, y_t)$, so that we have 
\begin{equation}
\label{eq:m}
\Phi_i(x_{t}, y_{t}) \approx {\gamma}_{t, i}, \hspace{0.2cm} \text{if} \; y_t = j
\end{equation}
with $i = (d-1)j + r$ for $j = 1, 2, ..., |\mathcal{Y}|$ and $r = 1, 2, ..., d$. Then, equations~\eqref{eq:e} and~\eqref{eq:m} above lead to the dynamical system in~\eqref{eq:kalman}.

\newpage
 \section{Implementation details and additional results} \label{app:imp_det}

 In this section we describe the implementation details used for the numerical results in Section~\ref{sec:er} and show additional results. In addition, the Python and Matlab implementations of the proposed \acsp{AMRC} are available on the web \url{https://github.com/MachineLearningBCAM/AMRC-for-concept-drift-ICML-2022} with the setting used in the numerical results. 

 In the synthetic dataset described in Section~\ref{sec:er}, the class-conditional underlying distributions $\up{p}_t(\V{x}_t|y_t = 1)$ and $\up{p}_t(\V{x}_t|y_t = 2)$ are Gaussian with means that move with varying velocity and direction in a circle centered at the origin. Specifically, the speed of such means has periodicity $\pi / \omega$ with maxima at times $t = ((n-1) \pi + \pi /2)/\omega$ and minima at times $t = n \pi/\omega$, for $n \in \mathbb{N}$. In addition, the direction of movement changes with the same periodicity at times when the velocity is minimum. 
 
 In the real-world datasets used in Section~\ref{sec:er}, we use $13$ public datasets that have been often used as benchmark in adaptive settings. These datasets are further described in Table~\ref{tab:datasets} that shows the number of instance-label pairs, the dimensionality of instances, and the number of labels. Three of such datasets  are large-scale (``Airlines'', ``Credit card'', and ``Poker''), four of them are medium-sized datasets  (``Weather'', ``Phishing'', ``Smart grid stability'', and ``Shuttle''), and the rest are short-sized datasets (``Elec2'', ``German'', ``Chess'', ``Usenet1'', ``Usenet2'', and ``Email''). For instance, the ``Airlines'' dataset used contains 539,383 instances with flight arrival and departure information and aims to predict if a flight will be delayed or not; the ``Weather'' dataset used contains 18,159 instances with daily measurements of weather factors and aims to predict if it will rain or not; and the ``Elec2'' dataset used contains 1,148 instances with  twice-daily measurements (12am and 12pm) of factors that affect load demand and price and aims to predict if the price will be higher or not. 
 
      \begin{table}
                          \captionsetup{labelfont={it}, labelsep=period, font=small}
\caption{Datasets information show number of instances, dimensionality of instances, and number of labels.}
                           \centering
\begin{tabular}{| c | r | r | r | r | r | r | r | r | r | r | r |}
\hline
Dataset & Time Steps & Dim. of instances & $|\mathcal{Y}|$  \\ \hline
{Weather} & 18,159 & 8 & 2 \\
{Elec2} & 1,148 & 4 & 2 \\
Airlines & 539,383 & 7 & 2 \\
{German} & 1,000 & 24 & 2 \\
Chess  & 503 & 8 & 2 \\
Usenet 1 & 1,500 & 99 & 2 \\
Usenet 2 & 1,500 & 99 & 2 \\
Email & 1,498 & 913 & 2 \\
C. card & 284,806 & 30  &2 \\
S. Grid & 60,000 & 13 & 2 \\
Shuttle & 100,001& 9 & 4 \\
{Poker} & 829,201 & 10 & 10 \\
\hline
\end{tabular}
\label{tab:datasets}
\end{table}

 The results in Table~\ref{tb:results} for methods that use kernels are obtained with a scaling parameter $\gamma$ calculated using a two-stage five-fold cross validation. Specifically, at the first stage, the values for the scaling factor are selected from $2^i$ for $i = \{-6, -3, 0, 3, 6\}$. At the second stage, if $\gamma_0 = 2^i$ where $i$ is the best parameter obtained at the first stage, then the values for the scaling parameters are selected from $ \gamma_0 2^i$, $i = \{-2, -1, 0, 1, 2\}$. The final value is $\gamma = \gamma_0 2^i$ where $i$ is the best parameter obtained at second stage. 
 
   \begin{figure}
        \centering
                           \begin{subfigure}{0.23\textwidth}
         \centering
\psfrag{Probability of error}[c][t][0.4]{Probability of error}
\psfrag{Order 0 method k = 0abc}[l][l][0.4]{$R(\up{h}_t)$ order $k = 0$}
\psfrag{Order 1}[l][l][0.5]{Order 1}
\psfrag{Order 2 method k = 2}[l][l][0.5]{$R(\up{h}_t)$ order $k = 2$}
\psfrag{Order 1 method k = 1abc}[l][l][0.4]{$R(\up{h}_t)$ order $k = 1$}
\psfrag{Upper boundabcdef}[l][l][0.4]{$R(\mathcal{U}_t)$ order $k = 0$}
\psfrag{Error}[c][t][0.4]{Error}
                  \psfrag{0}[][][0.3]{}
                  \psfrag{20}[][][0.3]{}
                  \psfrag{40}[][][0.3]{40}
                \psfrag{60}[][][0.3]{}
                \psfrag{80}[][][0.3]{80}
                \psfrag{100}[][][0.3]{}
                \psfrag{120}[][][0.3]{120}
                \psfrag{140}[][][0.3]{}
                \psfrag{0.15}[][][0.3]{0.15}
                \psfrag{0.25}[][][0.3]{0.25}
                \psfrag{0.35}[][][0.3]{0.35}
                \psfrag{0.45}[][][0.3]{0.45}
                \psfrag{0.2}[][][0.3]{0.2}
                \psfrag{0.3}[][][0.3]{0.3}
                \psfrag{0.4}[][][0.3]{0.4}
                \psfrag{0.5}[][][0.3]{0.5}
  \includegraphics[width=\textwidth]{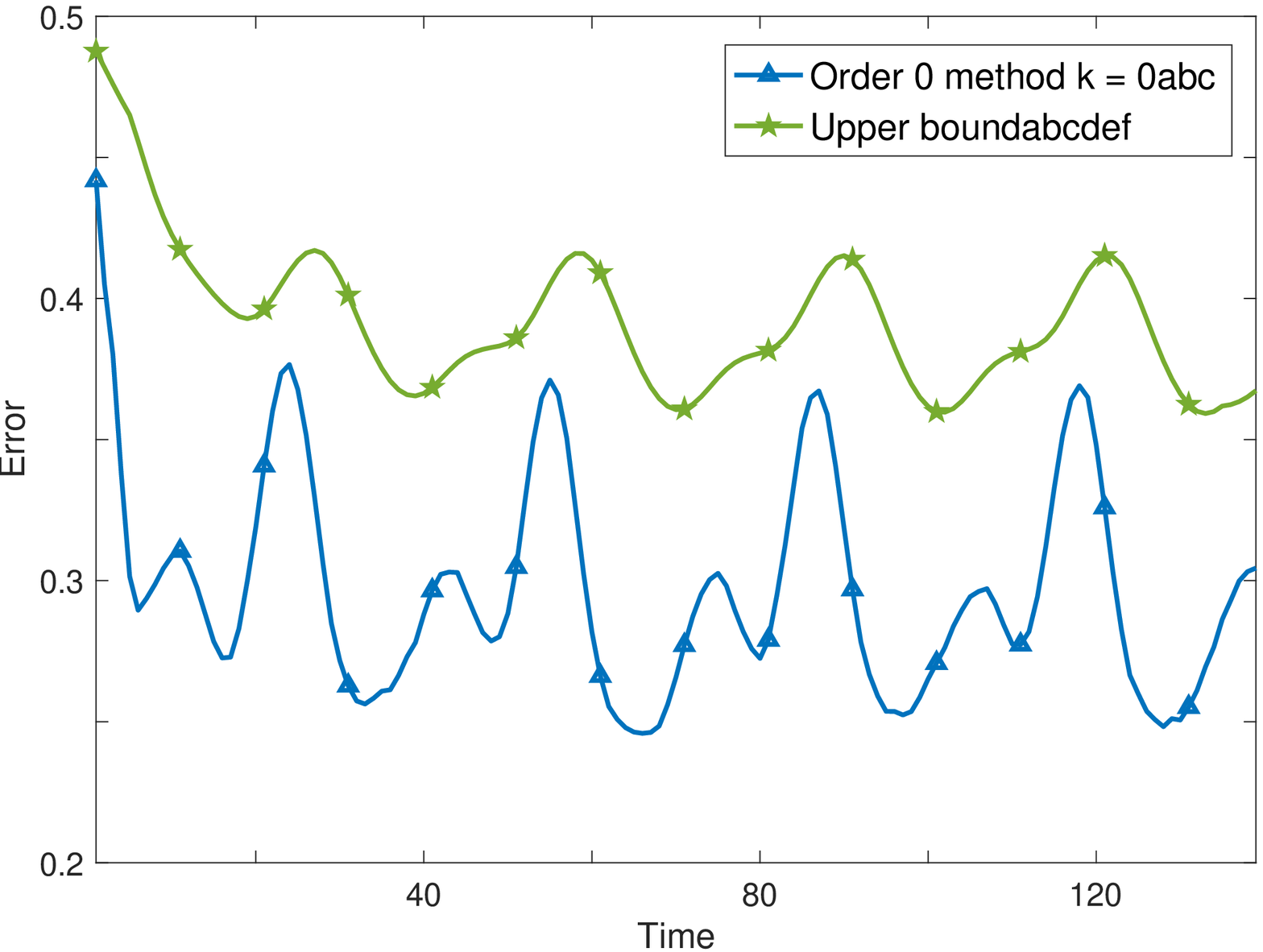}
   \captionsetup{labelfont={it}, labelsep=period, font=small}
                     \caption{{Instantaneous bounds and probabilities of error for AMRC of order $k = 0$.}}
         \label{fig:uppers0}
                  \end{subfigure}
                  \hfill
                                             \begin{subfigure}{0.23\textwidth}
         \centering
                           \psfrag{Time}[c][b][0.4]{Time $t$}
\psfrag{Probability of error}[c][t][0.5]{Probability of error}
\psfrag{Order 0 method k = 0abc}[l][l][0.5]{$R(\up{h}_t)$ order $k = 0$}
\psfrag{Order 1}[l][l][0.5]{Order 1}
\psfrag{Order 2 method k = 2abc}[l][l][0.4]{$R(\up{h}_t)$ order $k = 2$}
\psfrag{Order 1 method k = 1abc}[l][l][0.4]{$R(\up{h}_t)$ order $k = 1$}
\psfrag{Upper boundabcdef}[l][l][0.4]{$R(\mathcal{U}_t)$ order $k = 2$}
\psfrag{Error}[c][t][0.4]{Error}
                  \psfrag{0}[][][0.3]{}
                  \psfrag{20}[][][0.3]{}
                  \psfrag{40}[][][0.3]{40}
                \psfrag{60}[][][0.3]{}
                \psfrag{80}[][][0.3]{80}
                \psfrag{100}[][][0.3]{}
                \psfrag{120}[][][0.3]{120}
                \psfrag{140}[][][0.3]{}
                \psfrag{0.15}[][][0.3]{0.15}
                \psfrag{0.25}[][][0.3]{0.25}
                \psfrag{0.35}[][][0.3]{0.35}
                \psfrag{0.45}[][][0.3]{0.45}
                \psfrag{0.2}[][][0.3]{0.2}
                \psfrag{0.3}[][][0.3]{0.3}
                \psfrag{0.4}[][][0.3]{0.4}
                \psfrag{0.5}[][][0.3]{0.5}
          \includegraphics[width=\textwidth]{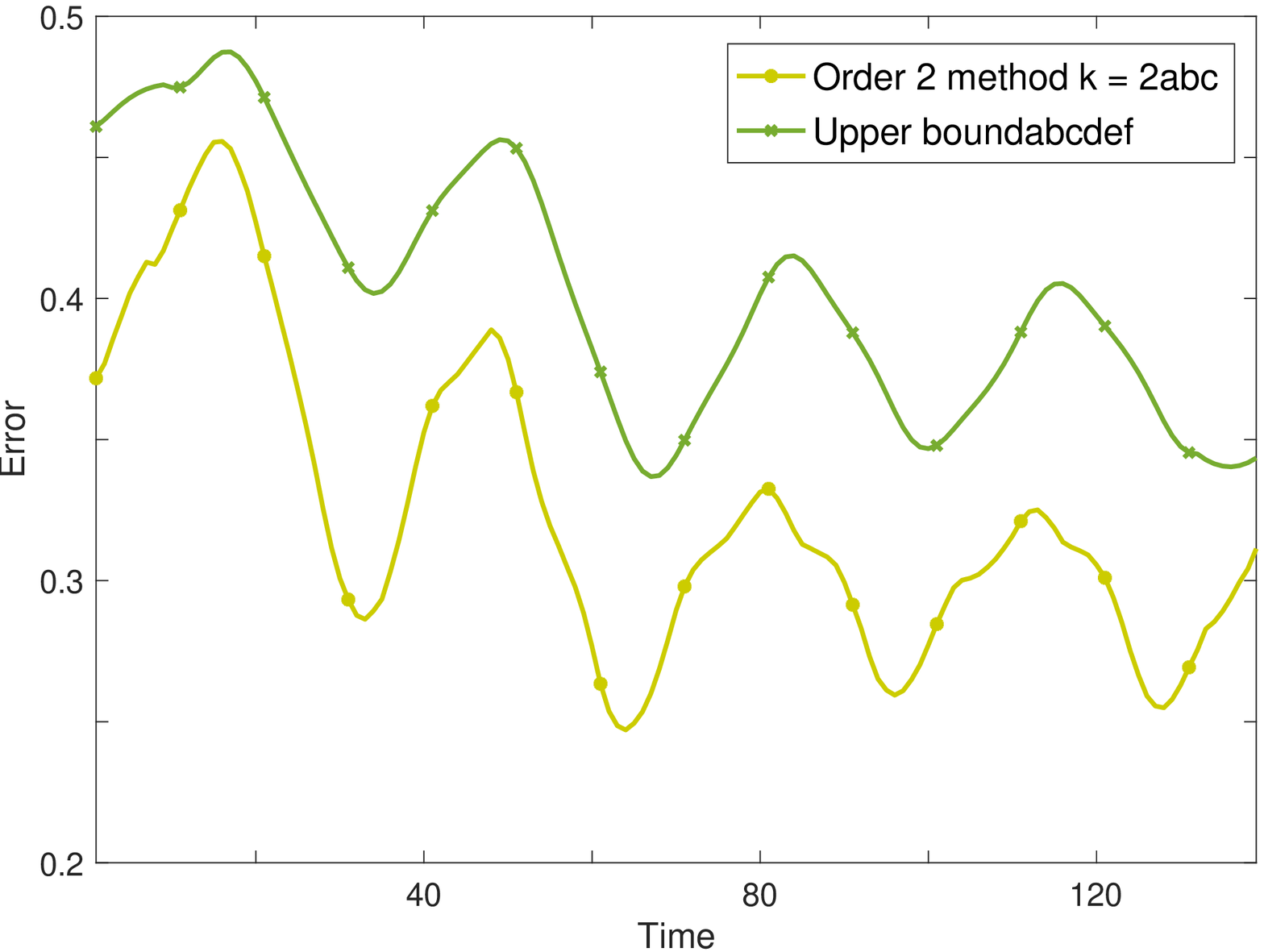}
           \captionsetup{labelfont={it}, labelsep=period, font=small}
                     \caption{{Instantaneous bounds and probabilities of error for AMRC of order $k = 2$.}}
         \label{fig:uppers2}
                  \end{subfigure}
                   \captionsetup{labelfont={it}, labelsep=period, font=small}
                                                 \caption{Results on synthetic data show the evolution in time of instantaneous bounds and probabilities of error.}          
\end{figure}

\begin{table}
\centering
\captionsetup{labelfont={it}, labelsep=period, font=small}
\caption{Averaged running times per time step in milliseconds of AMRC in comparison with state-of-the-art techniques.}
\label{tb:time}
       \footnotesize
\begin{tabular}{|c|r|}
\hline
\multicolumn{1}{|c|}{Algorithm} & Mean     \\ \hline
RBP                           & 9.2     \\ 
Proj.                         & 9.0          \\ 
P. ++                         & 9.1       \\ 
NORMA                         & 6.8    \\ 
A.Raker                       & 10.1     \\ 
NOGD                          & 4.6          \\ 
FOGD                          & 1.9 \\
$\lambda$-per.                & 12.2                  \\  
DWM                           & 6.6                 \\ 
Forg.                         & 6.9             \\ 
AMRC       $K = 500$    & 14.4    \\
AMRC $K = 2000$ & 51.9\\ \hline
\end{tabular}
\end{table}

In the first set of additional results we further illustrate the reliability of instantaneous bounds using the synthetic dataset described in Section~\ref{sec:er}. Specifically, we extend such results for \acsp{AMRC} with orders $k = 0$ and $2$ completing those in the main paper. In order to quantify the error probability at each time, we use 10,000 Monte Carlo simulations. Figures~\ref{fig:uppers0} and~\ref{fig:uppers2} show the averaged instantaneous bounds of probabilities of error in comparison with the true probabilities of error $R(\up{h}_t)$ at each time for \acsp{AMRC} with order $k = 0$ and $k = 2$, respectively. Such figures show that the instantaneous bounds $R(\mathcal{U}_t)$ obtained at \acsp{AMRC} learning can offer accurate estimates for the probability of error at each time.

In the second set of additional results, we assess the running time of AMRCs. Table~\ref{tb:time} shows the averaged running time  in milliseconds of AMRCs and the 10 state-of-the-art techniques for the 12 benchmark datasets. Such table shows the running time of AMRCs using $K = 500$ and $K = 2000$ iterations in the \ac{ASM} in Section~\ref{sec:lear}, and Table~\ref{tb:time2} shows the classification error of AMRCs with $K = 500$ and $K = 2000$ completing those in the main paper. The running time of AMRCs is of the order of tens of milliseconds per time step similarly to state-of-the-art techniques. 

\begin{table}
\centering
 \captionsetup{labelfont={it}, labelsep=period, font=small}
\caption{Classification error in $\%$ of AMRC with $K = 500$ and $K = 2000$ iterations in \ac{ASM}.}
\label{tb:time2}
       \footnotesize
\begin{tabular}{|c|r|r|r|r|}
\hline
\multirow{2}{*}{Dataset} & \multicolumn{2}{c|}{AMRC}    \\ 
& \multicolumn{1}{c}{$K = 500$} & \multicolumn{1}{c|}{$K = 2000$} \\\hline
Weather & 32.4&32.3 \\ 
Elec2 & 35.9 &35.8\\ 
Airlines & 39.2 &38.9\\ 
German& 30.4 &30.3 \\ 
Chess & 28.8& 27.7\\   
Usenet1  & 36.8 &35.7\\
Usenet2  & 30.6 &30.9\\ 
Email  & 43.0&43.7\\ 
C. card  & 2.53 & 0.17\\ 
S. grid  &36.3&35.8\\  
Shuttle  &44.6 &15.2\\ 
Poker  & 30.4 &21.9 \\ \hline
\end{tabular}
\end{table}

\end{document}